\pdfoutput=1
\documentclass[11pt]{article}

\usepackage[final]{acl}

\usepackage{times}
\usepackage{latexsym}

\usepackage[T1]{fontenc}
\usepackage[utf8]{inputenc}

\usepackage{microtype}

\usepackage{inconsolata}

\usepackage{graphicx} %
\usepackage{natbib}  %
\usepackage{caption} %
\usepackage{algorithm}
\usepackage{listings}
\usepackage{algorithmic}
\usepackage{amsmath, amsfonts, amssymb}
\usepackage{booktabs}
\usepackage{multirow}
\usepackage[outdir=./]{epstopdf}
\usepackage{enumitem}
\usepackage{caption}
\usepackage{subcaption}
\usepackage{subfloat}
\usepackage{newfloat}
\usepackage{graphicx}
\usepackage{svg} % svg
\usepackage[normalem]{ulem}
\usepackage{framed}
\usepackage{mdframed}
\usepackage{xcolor}
\usepackage{lipsum}
\usepackage{float}
\usepackage{hyperref}

\definecolor{shadecolor}{gray}{0.9}

\newlist{todolist}{itemize}{2}
\setlist[todolist]{label=$\square$}
\usepackage{pifont}

\usepackage{array}
\newcolumntype{L}[1]{>{\raggedright\let\newline\\\arraybackslash\hspace{0pt}}m{#1}}
\newcolumntype{C}[1]{>{\centering\let\newline  \\\arraybackslash\hspace{0pt}}m{#1}}
\newcolumntype{R}[1]{>{\raggedleft\let\newline \\\arraybackslash\hspace{0pt}}m{#1}}

\AtBeginDocument{%
  \providecommand\BibTeX{{%
    \normalfont B\kern-0.5em{\scshape i\kern-0.25em b}\kern-0.8em\TeX}}
}

\usepackage{authblk}

\title{PLPHP: Per-Layer Per-Head Vision Token Pruning for Efficient Large Vision-Language Models}

\author[ ]{Yu Meng$^{1}$\thanks{Equal contribution.}}
\author[1*]{Kaiyuan Li}
\author[1,3]{Chenran Huang}
\author[2]{Chen Gao}
\author[1]{Xinlei Chen}
\author[2]{Yong Li}
\author[1]{Xiaoping Zhang}

\affil[1]{Shenzhen International Graduate School, Tsinghua University}
\affil[2]{Tsinghua University \quad $^3$Tongji University}

\begin{document}
\maketitle
\begin{abstract}
Large Vision-Language Models (LVLMs) have demonstrated remarkable capabilities across a range of multimodal tasks. However, their inference efficiency is constrained by the large number of visual tokens processed during decoding. To address this challenge, we propose \textbf{P}er-\textbf{L}ayer \textbf{P}er-\textbf{H}ead Vision Token \textbf{P}runing (\textbf{PLPHP}), a two-level fine-grained pruning method including Layer-Level Retention Rate Allocation and Head-Level Vision Token Pruning. Motivated by the \textit{Vision Token Re-attention} phenomenon across decoder layers, we dynamically adjust token retention rates layer by layer. Layers that exhibit stronger attention to visual information preserve more vision tokens, while layers with lower vision attention are aggressively pruned. Furthermore, PLPHP applies pruning at the attention head level, enabling different heads within the same layer to independently retain critical context. Experiments on multiple benchmarks demonstrate that PLPHP delivers an 18\% faster decoding speed and reduces the Key-Value Cache (KV Cache) size by over 50\%, all at the cost of 0.46\% average performance drop, while also achieving notable performance improvements in multi-image tasks. These results highlight the effectiveness of fine-grained token pruning and contribute to advancing the efficiency and scalability of LVLMs. Our source code will be made publicly available.
\end{abstract}

\section{Introduction}
\label{sec::intro}

Recent advancements in Large Vision-Language Models (LVLMs) have established them as a prominent research focus in multimodal learning. Numerous open-source implementations have demonstrated remarkable capabilities across various tasks, including multimodal understanding and reasoning.

\begin{figure}[h!]
	\centering
	\subfloat[LLaVA-OneVision]{
		\includegraphics[width=0.23\textwidth]{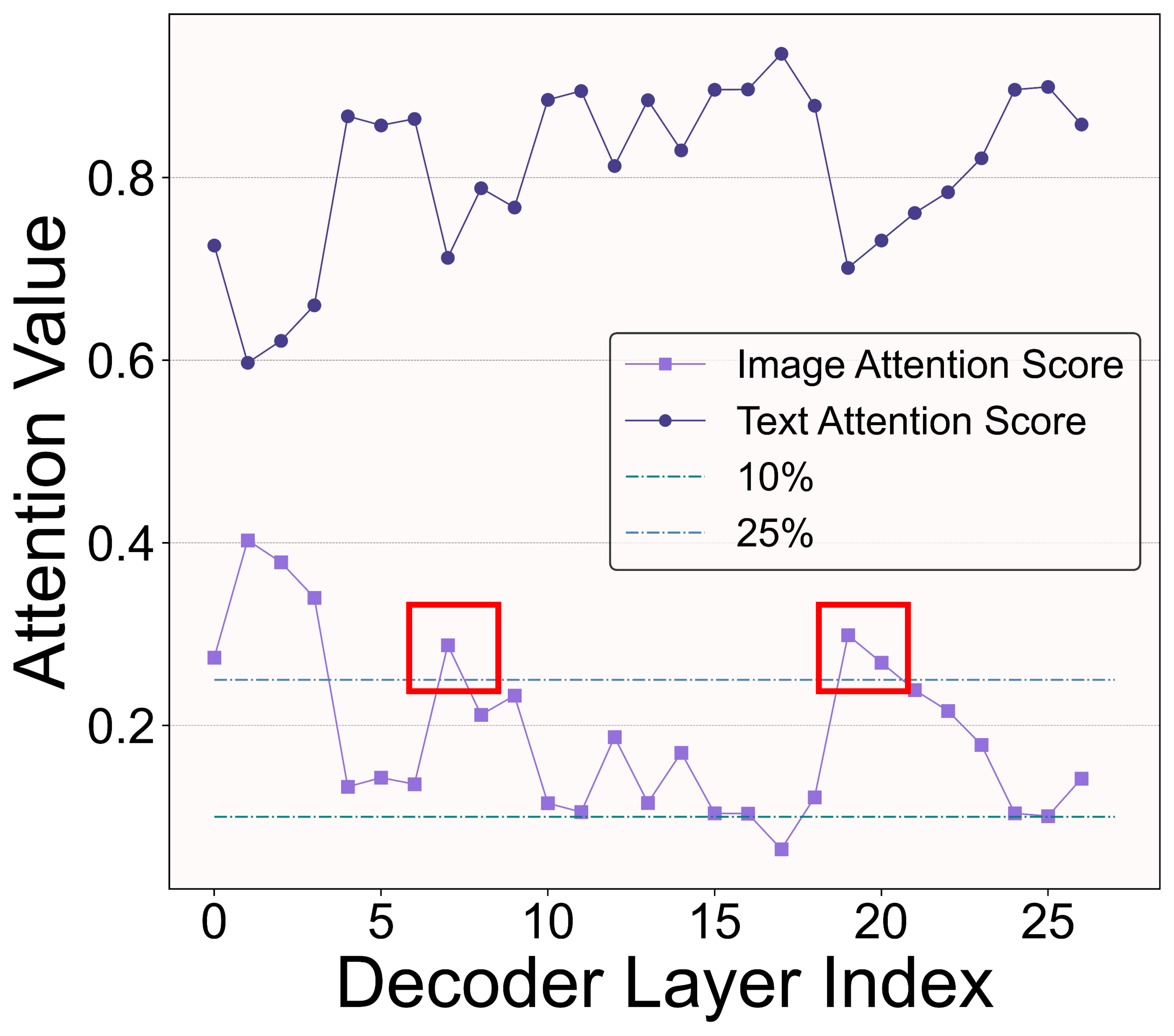}}
	\subfloat[Qwen2-VL]{
		\includegraphics[width=0.23\textwidth]{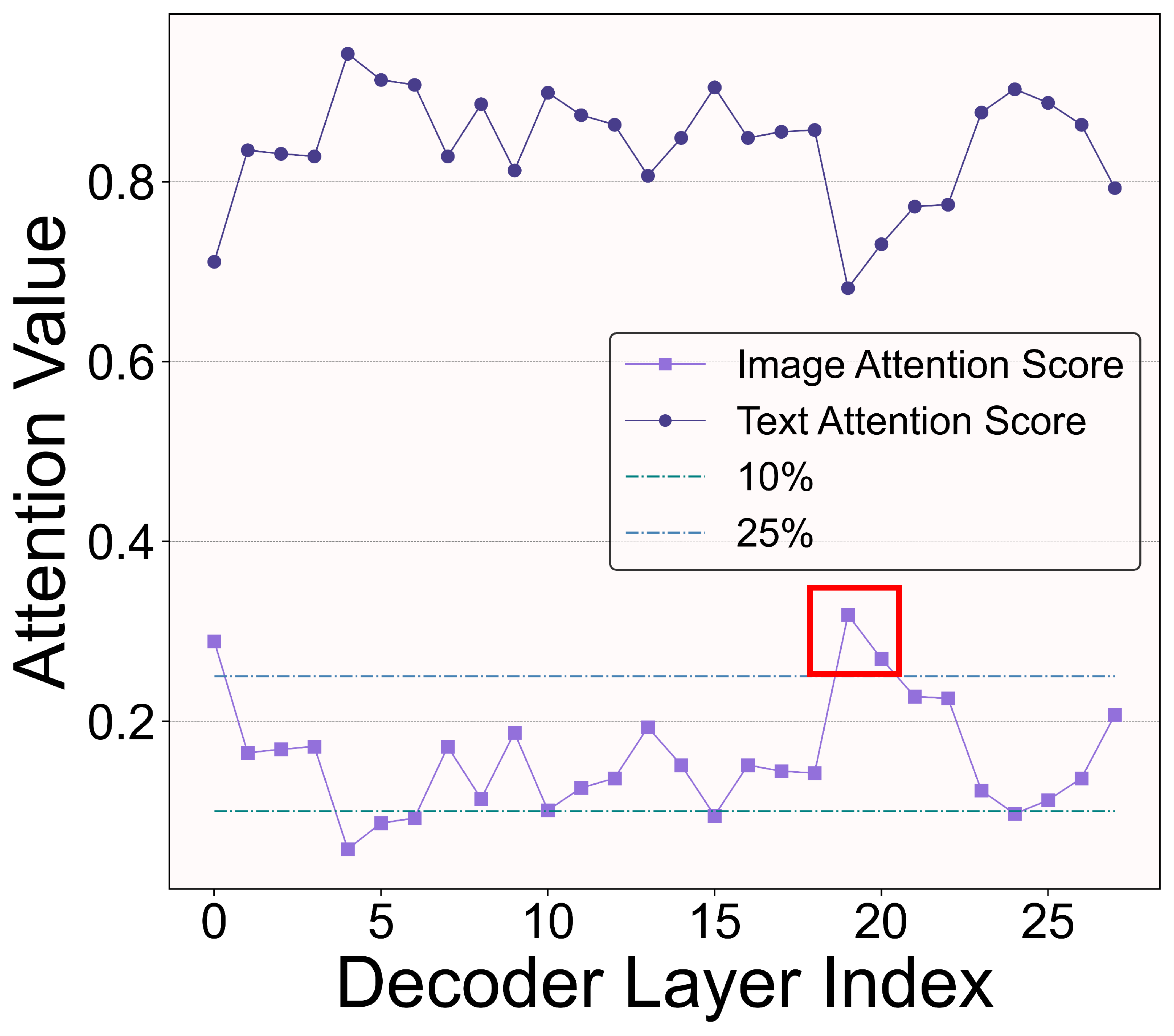}}
	\\ \quad \\ \quad
	\subfloat[IDEFICS2]{
		\includegraphics[width=0.23\textwidth]{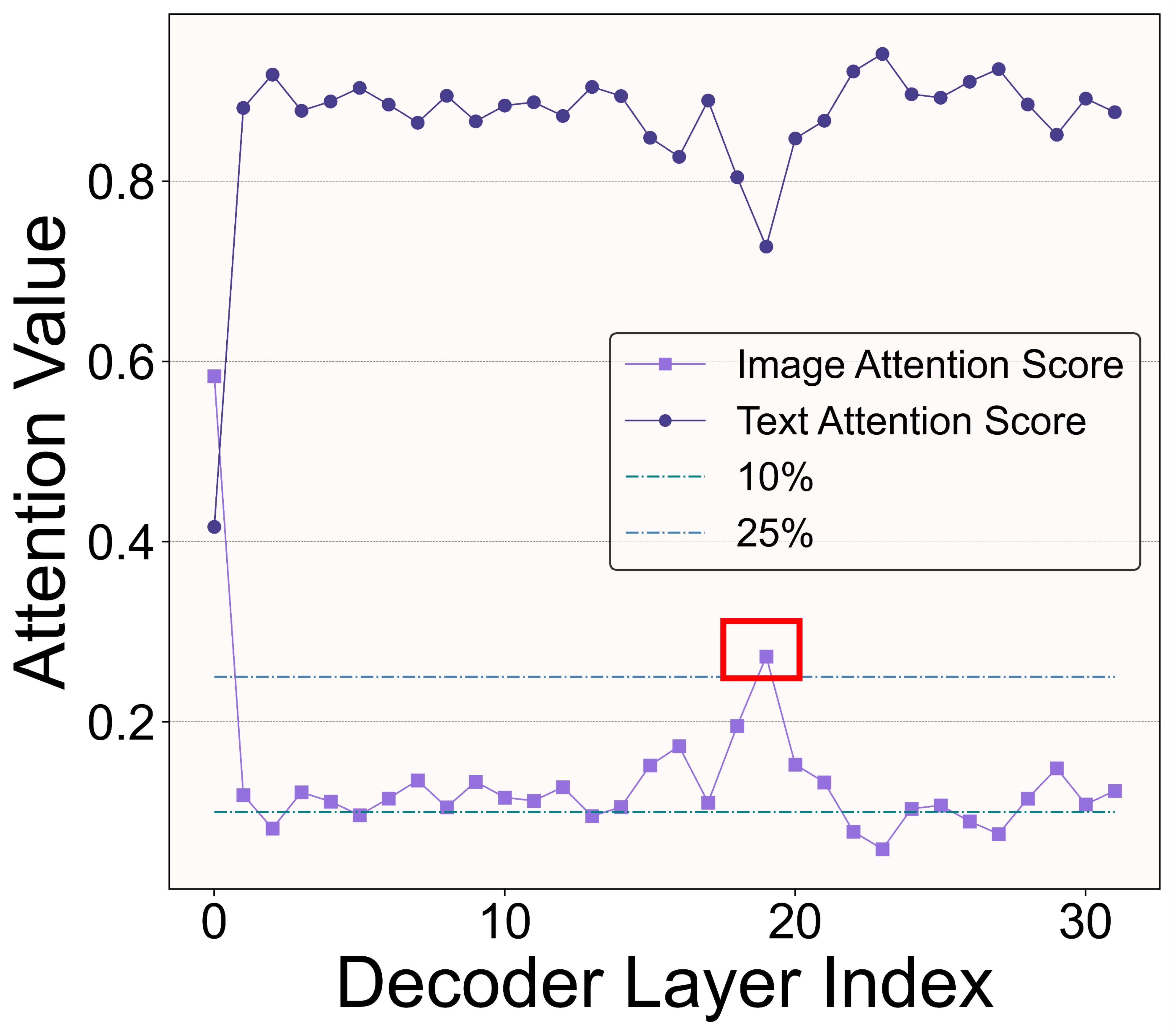}}
	\subfloat[Mantis]{
		\includegraphics[width=0.23\textwidth]{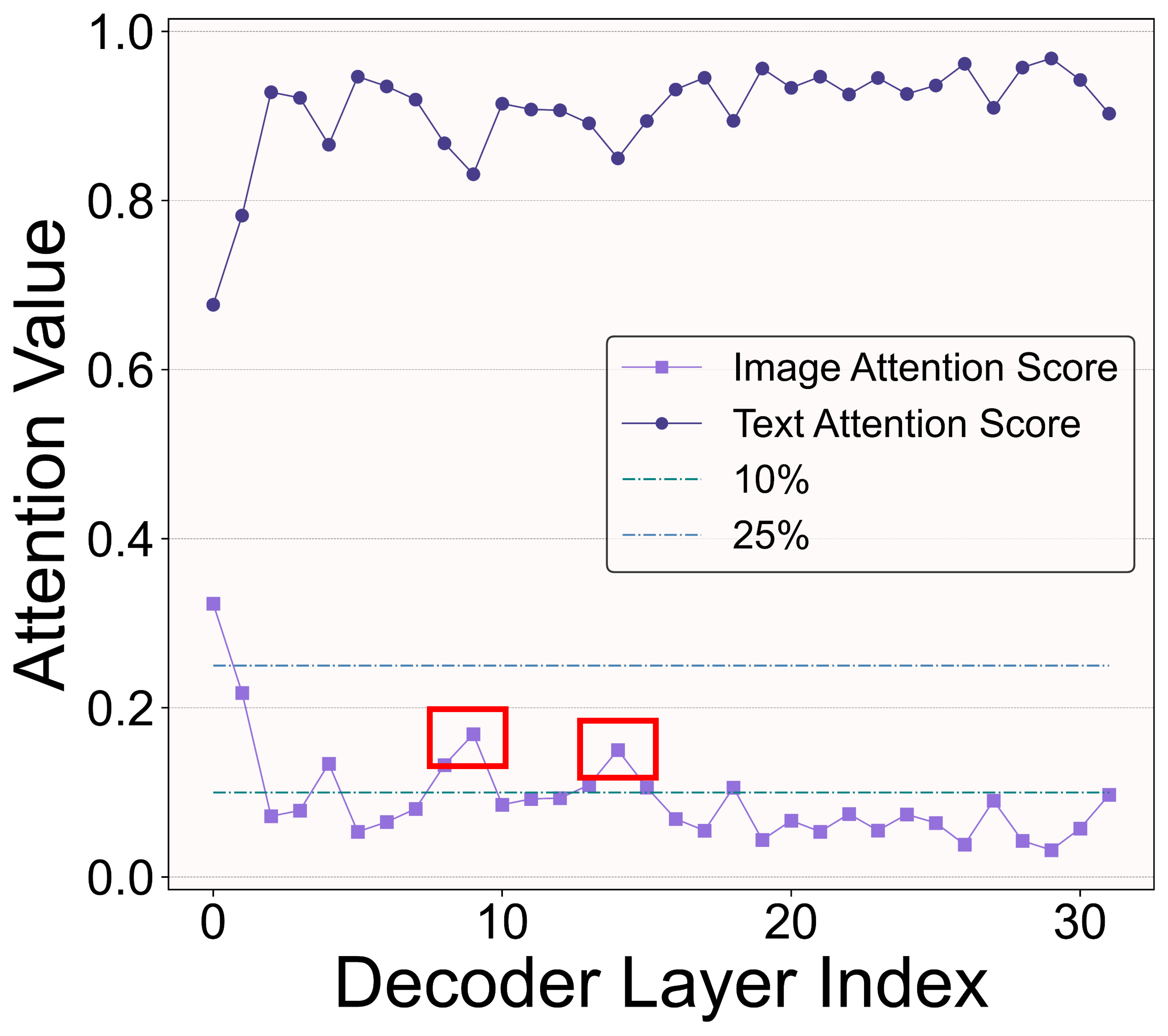}}
	\caption{\textbf{The phenomenon of Vision Token Re-attention in different LVLMs.} Various LVLMs demonstrate the phenomenon of refocusing on images within deep decoder layers. In these layers, the attention scores corresponding to vision tokens increase, as indicated by the \textcolor{red}{red boxes} highlighted in the figure.}
        \vspace{-0.4cm}
		\label{fig:refocus}
\end{figure}

Nevertheless, LVLMs face computational inefficiency challenges, mainly due to converting visual inputs into lengthy vision token sequences, ranging from thousands to tens of thousands. Previous studies \cite{chen2024image,lin2024boosting} find that LVLMs exhibit lower attentions to vision tokens in deeper layers compared to shallower layers, thus a certain amount of vision tokens are pruned at specific shallow layers, and the \textit{same} tokens are pruned in \textit{all} subsequent layers. However, such coarse-grained pruning strategies often lead to a significant performance decline in complex tasks that require comprehensive visual information, including open-ended VQA and image captioning. To address this challenge, in this work, we propose \textbf{P}er-\textbf{L}ayer \textbf{P}er-\textbf{H}ead Vision Token \textbf{P}runing (\textbf{PLPHP}), a plug-and-play adaptive fine-grained vision token pruning method that includes two levels: \textbf{1) Layer-Level Retention Rate Allocation} and \textbf{2) Head-Level Vision Token Pruning}, significantly reducing the performance loss associated with pruning.

\begin{figure}[h!]
	\centering
        \subfloat[LLaVA-OneVision]{
		  \includegraphics[width=0.23\textwidth]{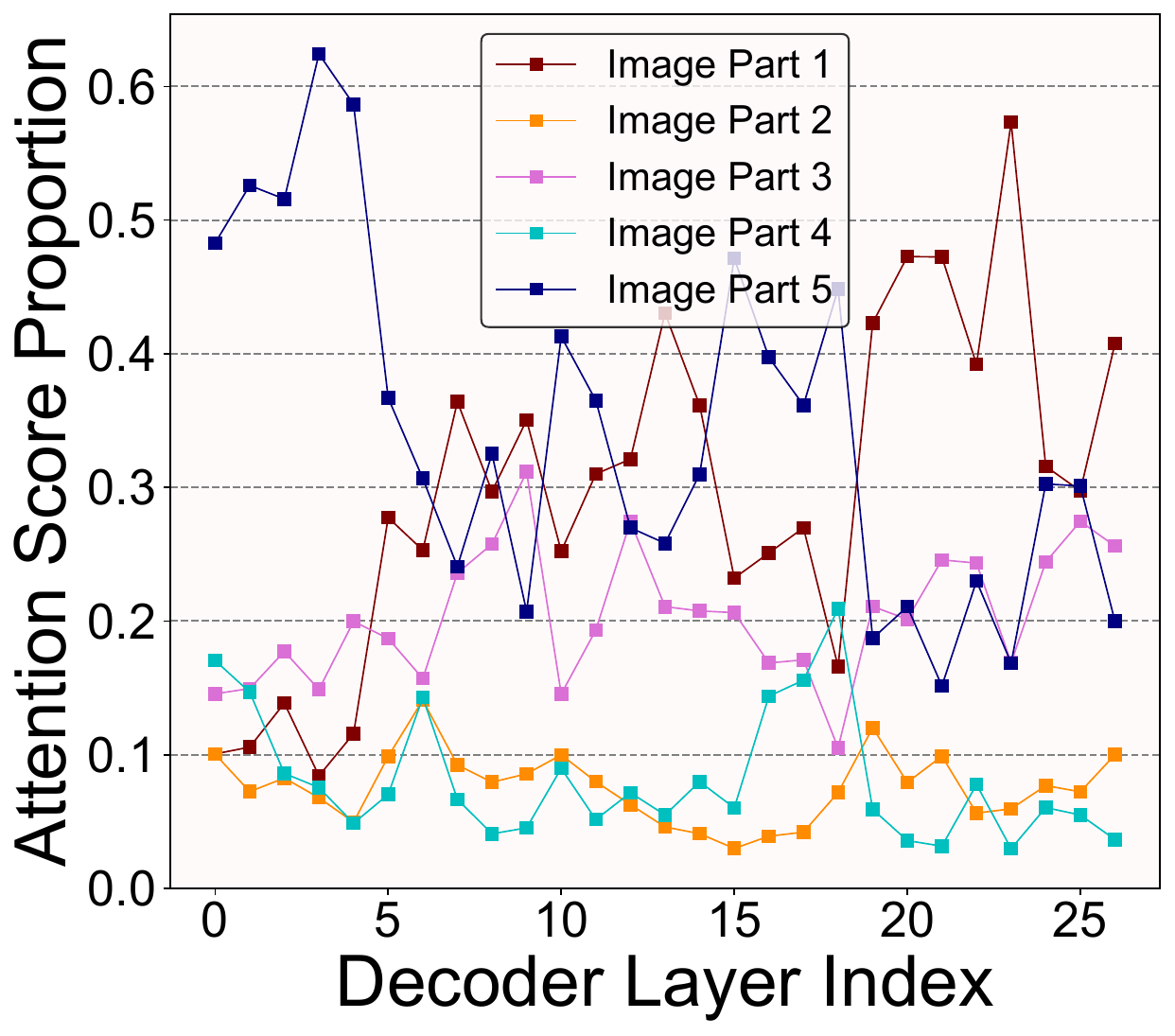}
        }
        \subfloat[Qwen2-VL]{
		  \includegraphics[width=0.23\textwidth]{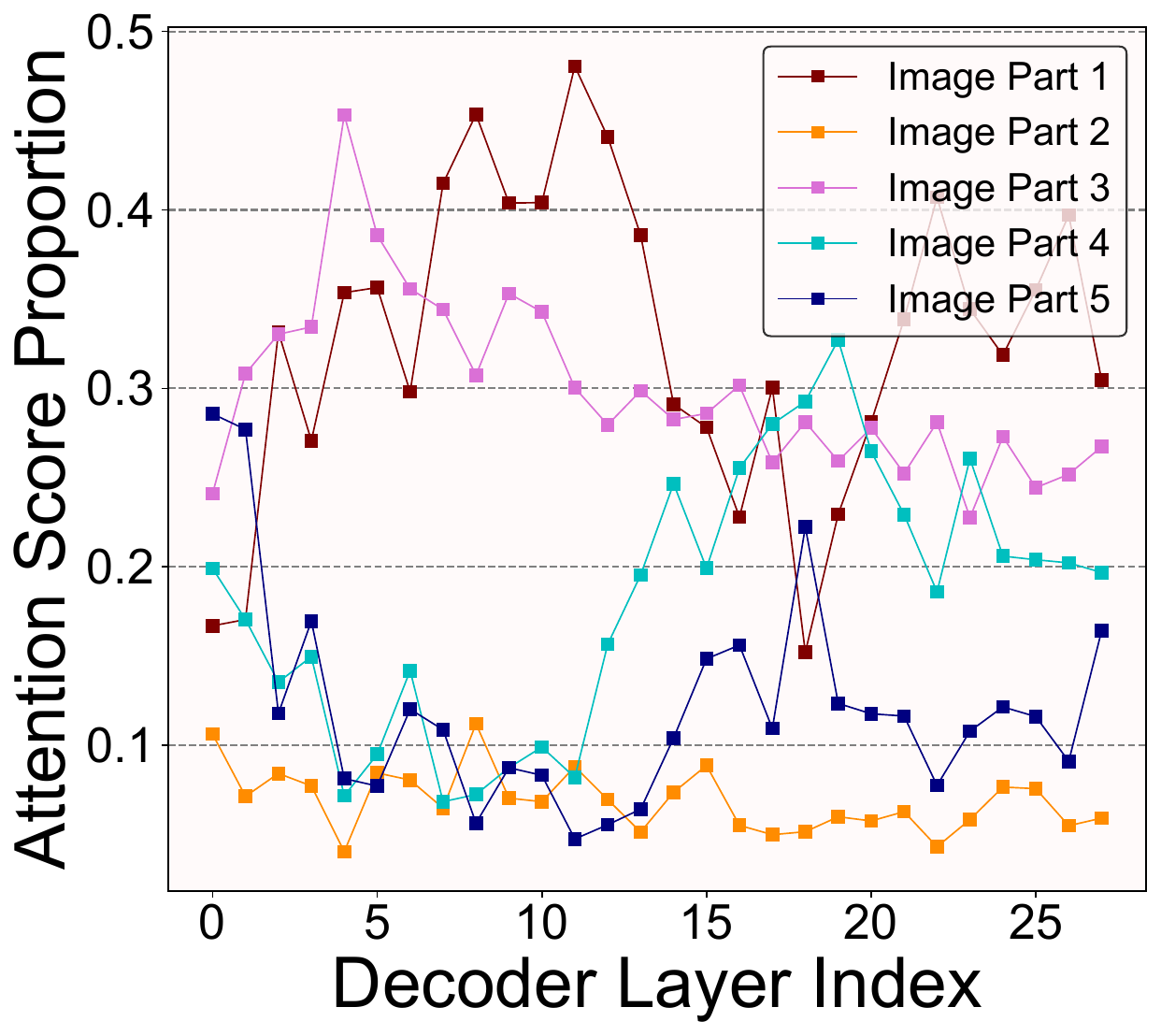}
        }
	\caption{\textbf{The proportion of attention scores received by different parts of the same image varies across different decoder layers.} Each polyline in the figure represents the proportion of attention scores for the corresponding group of tokens across different decoder layers.}
        \vspace{-0.2cm}
		\label{fig:image-part}
\end{figure}

The first level of our proposed method stems from our analysis of the attention to visual information in the deeper layers of LVLMs. As shown in Figure \ref{fig:refocus}, we observe the phenomenon of \textit{Vision Token Re-attention} across LVLMs with different architectures where attention scores of vision tokens are initially high and decrease in intermediate layers, but rise again in certain deeper layers. This indicates that LVLMs \textit{do not} always disregard vision tokens in deep layers, thus we need to dynamically adjust the pruning rate to accommodate the unique attention patterns of different decoder layers.

\begin{figure}[h!]
	\centering
	\subfloat[Head 2]{
		\includegraphics[width=0.16\textwidth]{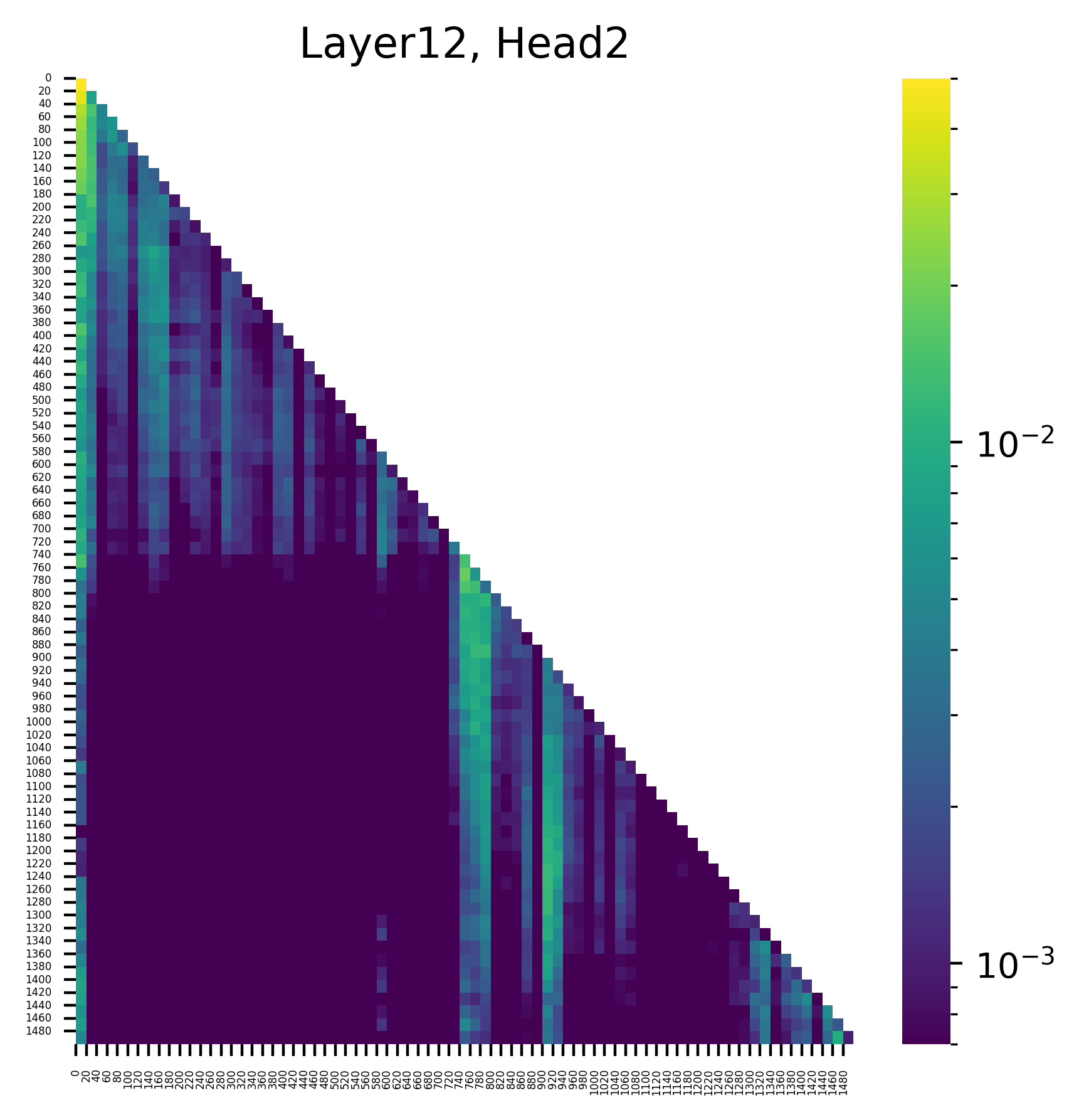}}
	\subfloat[Head 9]{
		\includegraphics[width=0.16\textwidth]{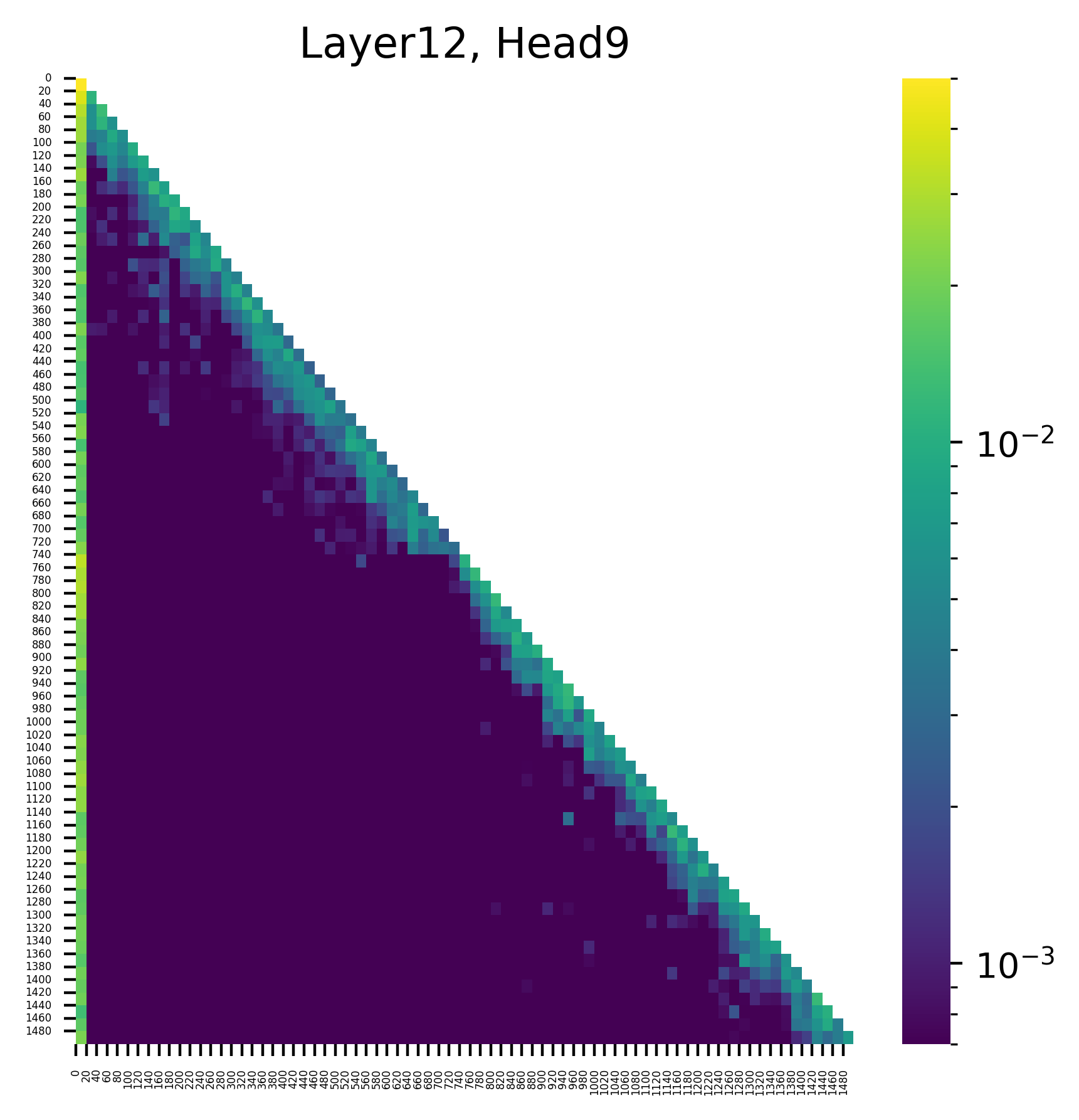}}
	\subfloat[Head 12]{
		\includegraphics[width=0.16\textwidth]{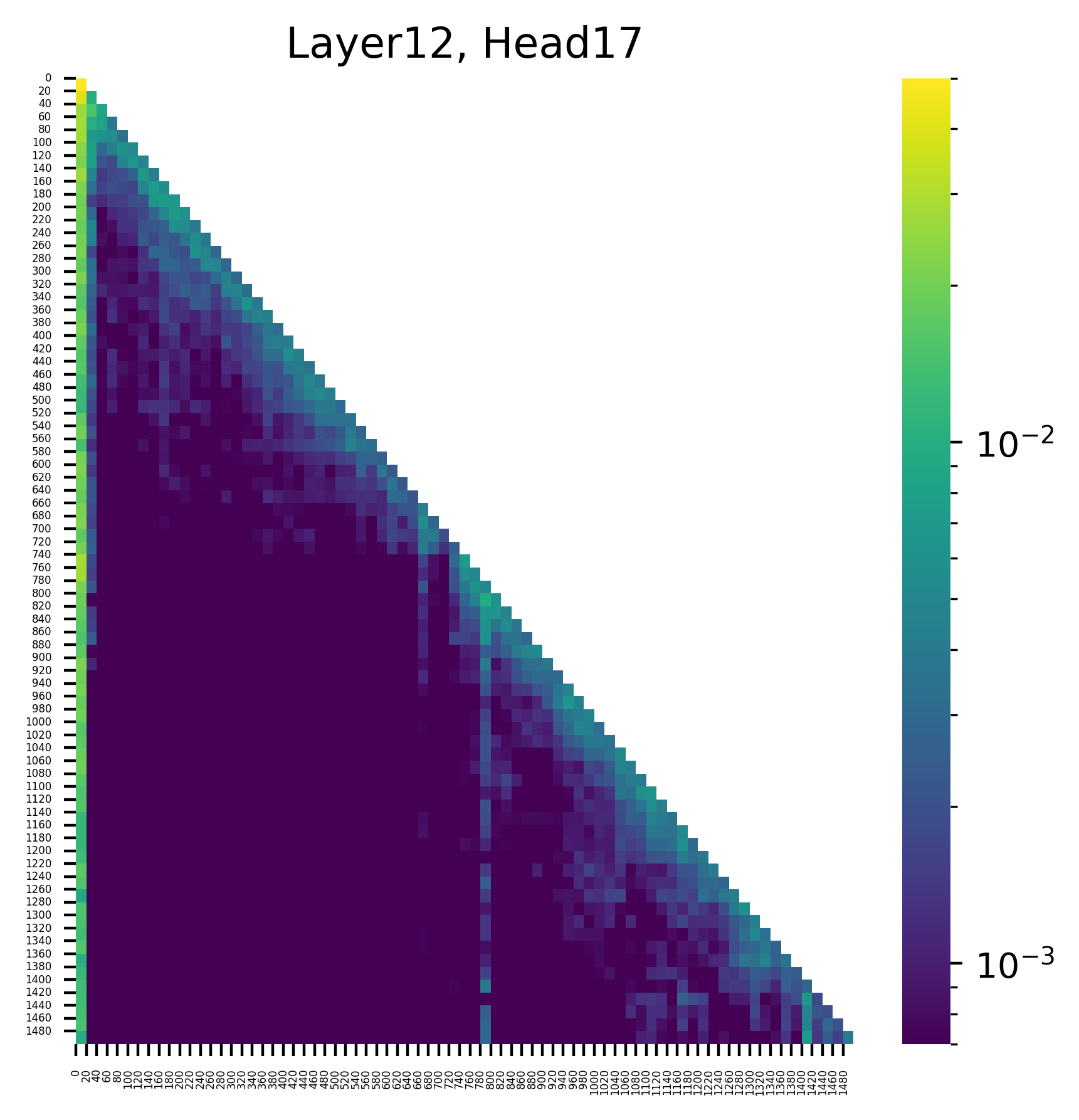}}
	\caption{\textbf{Visualization of attention maps in various attention heads.} Different heads within the same decoder layer exhibit different attention patterns.}
        \vspace{-0.2cm}
		\label{fig:per-head-attention-pattern}
\end{figure}

The second level of our method is motivated by an in-depth investigation on the variations in vision token attention across different decoder layers. As shown in Figure \ref{fig:image-part}, we divide the vision tokens into five groups based on their spatial relationships and plot the proportions of attention scores for each group across different layers. We observe that different parts of the same input image receive varying proportions of attention across different decoder layers, suggesting that each decoder layer specializes in processing distinct contexts. Furthermore, we conduct a more granular analysis at the level of attention heads. As illustrated in Figure \ref{fig:per-head-attention-pattern}, different attention heads within the same decoder layer exhibit distinct patterns of attention, demonstrating that the focus on different contexts occurs at the attention head level. This observation suggests that the unique contextual information processed by each attention head should be independently preserved during the pruning process to maintain model performance.

Built on these motivations, by dynamically adjusting retention rates according to layer-specific attention patterns layer by layer, PLPHP retains more vision tokens in layers where image attention scores are high, while aggressively pruning layers with low attention scores. Additionally, through head-level independent context pruning, PLPHP preserves the most critical contextual information for each attention head, leading to performance improvements. Comprehensive evaluations across multiple model architectures and various benchmarks demonstrate the effectiveness of PLPHP. Our method achieves over 50\% compression of the KV cache, over 18\% decoding acceleration, and only a 0.46\% average performance degradation with notable improvements on multi-image tasks.

The contributions of our work can be summarized into the following three points:
\begin{itemize}[leftmargin=*]
\setlength{\topmargin}{0pt}
\setlength{\itemsep}{0em}
    \item  We uncover the widespread phenomenon of \textit{Vision Token Re-attention} through investigations on various LVLMs, which could be a significant factor leading to the performance degradation of existing pruning methods.
    \item  We propose PLPHP, a plug-and-play adaptive fine-grained vision token pruning method that improves the performance of pruned models significantly while maintaining high computational efficiency.
    \item We conduct extensive experiments across multiple benchmarks and model architectures, validating the superiority of our proposed method.
\end{itemize}
\section{Related Work}
\label{sec::related}

\subsection{Large Vision-Language Models}\label{sec::relatedwork1}
Recent advancements in LVLMs significantly enhanced multimodal content understanding. \citet{liu2023llava} developed LLaVA, an early general-purpose multimodal model integrating CLIP \cite{radford2021learningtransferablevisualmodels} with language models. Subsequent innovations include Qwen-VL \cite{bai2023qwen,wang2024qwen2}, which enhanced visual processing with a specialized visual receptor and multilingual corpus, and Mantis by \citet{jiang2024mantis}, which improved multi-image reasoning through academic-level instruction tuning. \citet{laurenccon2024obelics} introduced IDEFICS, trained on the OBELICS dataset of interleaved image-text documents. Unified approaches by \citet{li2024llava-interleave} and \citet{li2024llava-onevision} achieved state-of-the-art performance in single-image, multi-image, and video tasks. However, LVLMs still face computational challenges due to the high number of visual tokens during inference, underscoring the need for more efficient inference.

\subsection{Efficient Multimodal Large Language Models}\label{sec::relatedwork2}
To optimize the computational efficiency of LVLMs during inference, works such as MobileVLM \cite{chu2023mobilevlm}, Tinygpt-V \cite{yuan2023tinygpt}, MoE LLaVA \cite{lin2024moe}, and LLaVA-Phi \cite{zhu2024llava} proposed more efficient model architectures. Meanwhile, \citet{li2023distilling} introduced a model-distillation approach that transfers knowledge from large vision-language models (VLMs) to smaller, lighter counterparts. Q-VLM \cite{wang2024q} provided a post-training quantization framework for LVLMs by mining cross-layer dependencies to improve quantization efficiency. From the perspective of token pruning, TokenPacker \cite{li2024tokenpacker}, Dynamic-LLaVA \cite{huang2024dynamic}, and AVG-LLaVA \cite{lan2024avg} investigated training LVLMs with fewer vision tokens to boost computational efficiency. However, these methods typically require additional model training, which imposes further computational overhead.

Training-free token pruning has also been widely employed in prior research to alleviate token redundancy in vision transformers (ViTs) and large language models (LLMs). For example, PruMerge \cite{shang2024llava} and VisionZip \cite{yang2024visionzip} suggested strategies to reduce vision tokens generated by vision encoders, thereby lowering vision token volume. FastV \cite{chen2024image} and SparseVLM \cite{zhang2024sparsevlm} observed that visual tokens become less significant in deeper layers, thus proposing to eliminate redundant vision tokens during inference. VTW \cite{lin2024boosting} introduced a strategy to remove all vision tokens at a specific layer based on KL Divergence. Although these methods have demonstrated effectiveness, they overlook the distinctions among different layers and attention heads within LVLMs, leading to a significant performance decline on complex tasks. Our research addresses this gap by proposing a fine-grained pruning method including both Layer-Level Retention Rate Allocation and Head-Level Vision Token Pruning.

\section{Method}\label{sec::method}

Our method is a plug-and-play module during the inference process of LVLMs. Therefore, we first outline the inference process of LVLMs as preliminary, followed by the design of our proposed PLPHP.

\begin{figure*}
	\centering
		\includegraphics[width=\textwidth]{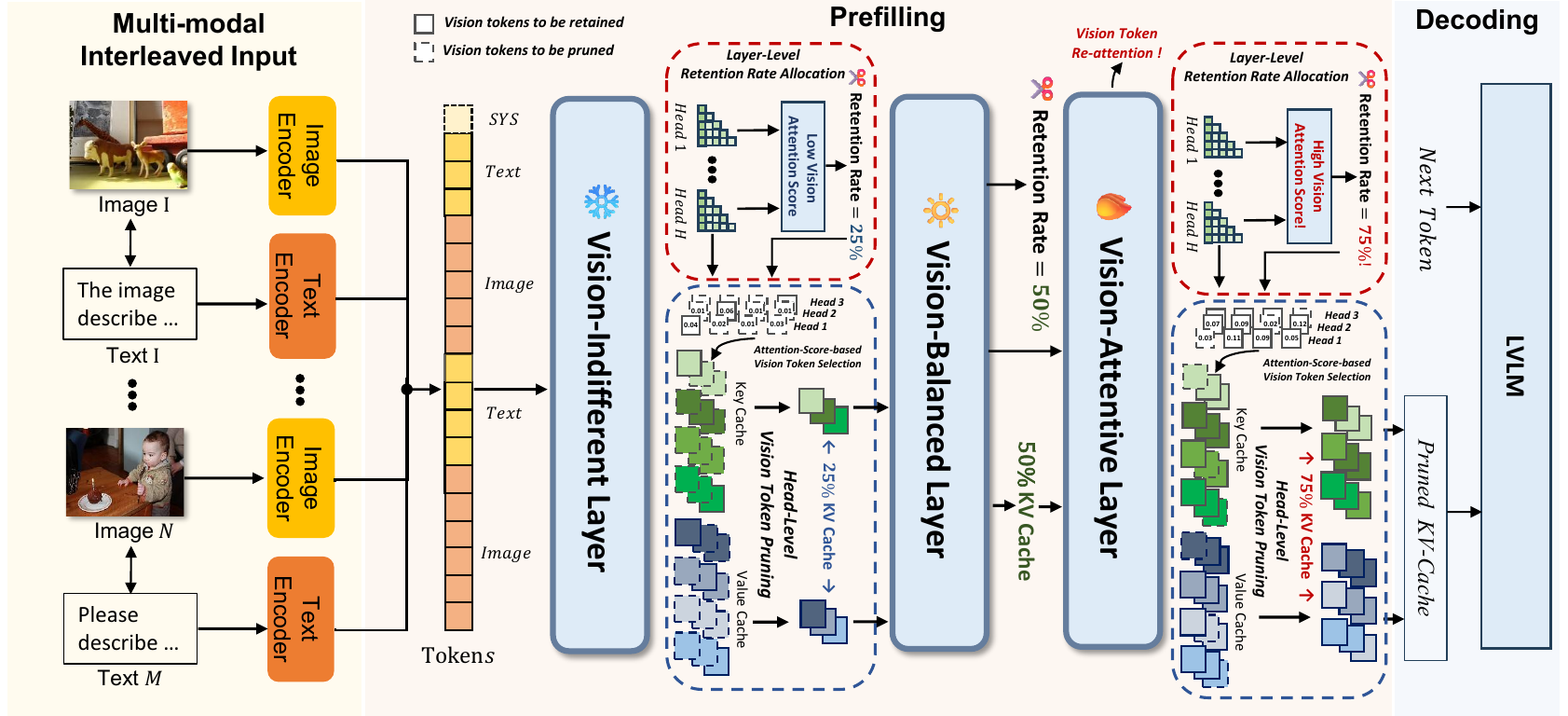}
	\caption{\textbf{Overview of PLPHP.} PLPHP has a two-level design including \textbf{Layer-Level Retention Rate Allocation} (as indicated by the \textcolor{red}{red dashed boxes}) and \textbf{Head-Level Vision Token Pruning} (as indicated by the \textcolor{blue}{blue dashed boxes}). Upon the completion of prefilling a certain decoder layer, PLPHP categorizes the layer as vision indifferent, balanced or attentive, and assigns a vision token retention rate to the layer based on its average attention scores to the vision tokens. Subsequently, according to the allocated retention rate, PLPHP performs fine-grained pruning for each head within the layer. It removes the visual tokens with lower attention scores from the KV cache of each attention head, ensuring that the remaining proportion of vision tokens does not exceed the pre-assigned retention rate.}
        \vspace{-0.5cm}
		\label{fig:arch}
\end{figure*}

\subsection{Preliminary}

LVLMs typically employ‌ an autoregressive generation paradigm during inference, which comprises two stages: the Prefilling Stage and the Decoding Stage.

\noindent \textbf{Prefilling Stage.}  In the Prefilling Stage, different modalities are mapped into a sequence of embedding vectors (tokens), which serves as the input to the LLM backbone. We denote the interleaved multimodal input token sequence of $m$ text segments and $n$ images $\mathbf{X}^1  \in \mathbb{R}^{S \times D}$ as:
\begin{equation}
\mathbf{X}^1 = \begin{pmatrix}
    \mathbf{X}_1^{(T)}\\ \mathbf{X}_1^{(I)}\\ \vdots, \\ \mathbf{X}_m^{(T)}\\ \mathbf{X}_n^{(I)}
\end{pmatrix},
\end{equation}
where $\mathbf{X}_i^{(T)} \in \mathbb{R}^{S_i^{(T)} \times D}$ represents the token sequence of the $i$-th text segment, and $\mathbf{X}_j^{(I)} \in \mathbb{R}^{S_j^{(I)} \times D}$ represents the token sequence of the $j$-th image. $S_i^{(T)}$ and $S_j^{(I)}$ represent the number of tokens for the $i$-th text segments and the $j$-th image, respectively, while \(S = \sum_{i=1}^m S_i^{(T)} + \sum_{j=1}^n S_j^{(I)}\) represents the total length of the input token sequence. $\mathcal{I}^{\left(T\right)}_i \in \mathbb{N}_0^{S^{(T)}_i}$ and $\mathcal{I}^{\left(I\right)}_j \in \mathbb{N}_0^{S^{(I)}_j}$ denote the corresponding token index sets of $\mathbf{X}_i^{\left(T\right)}$ and $\mathbf{X}_j^{\left(I\right)}$ within $\mathbf{X}^1$.

$\mathbf{X}^1$ is then fed into an LLM composed of $N$ decoder layers. Since the output and input shapes of each decoder layer are the same, we can denote the input of the $l$-th decoder layer as $\mathbf{X}^l \in \mathbb{R}^{S \times D}$. For the $h$-th attention head in the $l$-th layer:
\begin{equation}\label{eq:calcq}
\mathbf{Q}^{l, h} = \mathbf{X}^l \mathbf{W}_Q^{l, h},
\end{equation}
\begin{equation}\label{eq:calck}
\mathbf{K}^{l, h} = \mathbf{X}^l \mathbf{W}_K^{l, h},
\end{equation}
\begin{equation}\label{eq:calcv}
\mathbf{V}^{l, h} = \mathbf{X}^l \mathbf{W}_V^{l, h},
\end{equation}
where $\mathbf{W}_Q^{l, h} \in \mathbb{R}^{D \times D_k}$, $\mathbf{W}_K^{l, h} \in \mathbb{R}^{D \times D_k}$, and $\mathbf{W}_V^{l, h} \in \mathbb{R}^{D \times D_k}$ are referred to as the query projector, key projector, and value projector, respectively. $D_k$ is called the head dimension. $\mathbf{K}^{l, h}$ and $\mathbf{V}^{l, h}$ are then stored as the KV cache for the current attention head.

The attention weights $\mathbf{A}^{l,h} \in \mathbb{R}^{S\times S}$ are given by:

\vspace{-0.3cm}
\begin{equation}
    \mathbf{A}^{l,h} = \text{Softmax}\left(\frac{\mathbf{Q}^{l,h}\left(\mathbf{K}^{l,h}\right)^\top + \mathbf{\Lambda}}{\sqrt{D_k}}\right),
\end{equation}
where $\mathbf{\Lambda} \in \mathbb{R}^{S \times S}$ is an upper triangular matrix whose non-zero values are set to $-\inf$ and diagonal elements are set to $0$.

\noindent \textbf{Decoding Stage.} During the Decoding Stage, the model sequentially generates tokens and updates the KV cache of each attention head. At each timestep $t$, the input to the $l$-th decoder layer is a single token $\mathbf{x}^{l}_t \in \mathbb{R}^{1\times D}$. For the $h$-th attention head of the $l$-th layer, the KV cache is updated by:
\begin{equation}
    \mathbf{K}^{l,h} \leftarrow \begin{pmatrix} \mathbf{K}^{l,h} \\ \mathbf{x}^l_t\mathbf{W}^{l,h}_K\end{pmatrix},
\end{equation}
\begin{equation}
    \mathbf{V}^{l,h} \leftarrow \begin{pmatrix} \mathbf{V}^{l,h} \\ \mathbf{x}^l_t\mathbf{W}^{l,h}_V \end{pmatrix}. 
\end{equation}

\subsection{PLPHP}

\subsubsection{Overview}

PLPHP is a two-level adaptive fine-grained pruning method with \textbf{Layer-Level Retention Rate Allocation} and \textbf{Head-Level Vision Token Pruning}. The architecture is illustrated in Figure \ref{fig:arch}.

\subsubsection{Layer-Level Retention Rate Allocation}

To measure the extent of a decoder layer's attention to visual information, thereby determining the number of vision tokens to retain, we define the \textit{Vision Attention Score} $\gamma^l$ of the $l$-th layer as:
\begin{equation}
    \gamma^l = \sum_{k \in \bigcup_{j=1}^n\mathcal{I}^{\left(I\right)}_j}\frac{1}{H}\sum_{h=1}^H \mathbf{A}^{l,h}_{S,k},
\end{equation}
where $H$ represents the number of attention heads in each decoder layer. Note that the value of $\gamma^l$ is between $0$ and $1$. The larger the value of $\gamma^l$, the higher the $l$-th layer's attention to visual information.

In order to properly allocate the vision token retention rate based on the Vision Attention Score, given two thresholds $\alpha$ and $\beta$ ($0 \le \beta \le \alpha \le 1$), the $l$-th decoder layer is categorized as a \textbf{vision-attentive} layer when $\gamma^l \ge \alpha$, as a \textbf{vision-indifferent} layer if $\gamma^l < \beta$, and as a \textbf{vision-balanced} layer otherwise. The token retention rate $r^l$ for the $l$-th layer is defined as:
\begin{equation}
    r^l = \begin{cases}r + \Delta r, \quad &\gamma^l \ge \alpha \\ r - \Delta r,\quad &\gamma^l < \beta \\ r,\quad &\text{otherwise}\end{cases},
\end{equation}
where $0 \le \Delta r \le r \le 1 - \Delta r$. For example, selecting $\alpha=0.25$, $\beta=0.1$, $r=0.4$, and $\Delta r=0.3$ signifies that we regard decoder layers with $\gamma^l \ge 0.25$ as vision-attentive layers, and decoder layers with $\gamma^l < 0.1$ as vision-indifferent layers. For vision-attentive layers, we retain $0.4+0.3$, that is, $70\%$ of the vision tokens. For vision-indifferent layers, we retain $0.4-0.3$, that is, only $10\%$ of the visual tokens. For vision-balanced layers, we retain $40\%$ of the visual tokens. 

Through this dynamic calculation of token retention rates, we retain a larger number of vision tokens for the vision-attentive layers to leverage their heightened focus on image information, while we keep fewer vision tokens for the vision-indifferent layers to achieve higher efficiency with the least sacrifice of critical visual information. As for the vision-balanced layers, we strike a compromise, seeking an equilibrium between efficiency and performance.

\subsubsection{Head-Level Vision Token Pruning}

Given the retention rate $r^l$ calculated in the first level, we proceed to perform fine-grained pruning. According to FastV and \citet{zhang2024redundancy}, LVLMs typically exhibit a global focus on images in the first two layers and the last layer. Therefore, for a model composed of $N$ decoder layers, we select the third layer and the penultimate layer as the starting and ending layers for pruning.

To extract the most important vision tokens to preserve, for the $h$-th ($1 \le h \le H$) attention head in the $l$-th layer ($3 \le l \le N-1$), we calculate the indices of vision tokens with the highest attention scores within the $j$-th image input, accounting for the proportion $r^l$:
\begin{equation}
    \mathcal{I}^{\left(I_R\right), h}_{j} = \text{argtop}K_j\left(\mathbf{A}^{l,h}_{S}\left[\mathcal{I}_{j}^{\left(I\right)}\right]\right),
\end{equation}
where $K_j= r^l S_{j}^{\left(I\right)}$ and the $\text{argtop}K$ operation identifies the indices of the top $K$ elements with the highest values in the given sequence.

We then prune vision tokens by updating the key cache and value cache of the attention head by:
\begin{equation}\label{eq:update-k}
    \mathbf{K}^{l, h} \leftarrow \mathbf{K}^{l, h}\left[\bigcup_{i=1}^{m}\mathcal{I}^{\left(T\right)}_{i} \cup \bigcup_{j=1}^{n} \mathcal{I}_{j}^{\left(I_R\right), h} \right],
\end{equation}
\begin{equation}\label{eq:update-v}
    \mathbf{V}^{l, h} \leftarrow \mathbf{V}^{l, h}\left[\bigcup_{i=1}^{m}\mathcal{I}^{\left(T\right)}_{i} \cup \bigcup_{j=1}^{n} \mathcal{I}_{j}^{\left(I_R\right), h} \right],
\end{equation}
where $\left[\cdot\right]$ represents the indexing operation, which retrieves elements from a sequence according to the given indices.

To provide an intuitive explanation, for every attention head of the $l$-th decoder layer, we retain only the top $r^l$ proportion of the most attended tokens for each image, and remove the remaining $1-r^l$ proportion from the context. Since the number of text tokens is typically negligible compared to vision tokens, we retain all text tokens.

Our method allows different attention heads within the same decoder layer to selectively drop different contexts, thereby better utilizing the property of multi-head attention mechanisms where distinct heads can focus on various parts of the contextual information.

\section{Experiments}\label{sec::exp}

\begin{table*}[h!]
    \centering
    \caption{\textbf{Comparison of different methods on Multi-Image and Single-Image benchmarks.} $(\cdot)$ signifies the values by which the performance exceeds that of the uncompressed model after applying the corresponding method.}
    \renewcommand{\arraystretch}{1.1}
    \resizebox{\textwidth}{!}{
    \begin{tabular}{lccccccc}
        \toprule
        & \multicolumn{4}{c}{Multi-Image} & \multicolumn{3}{c}{Single-Image} \\
        \cmidrule(lr){2-5} \cmidrule(lr){6-8}
        & Spot-the-Diff & Image-Edit & Visual-Story-Telling & Multi-View & Flickr30k & COCO 2017 & DetailCaps4870 \\
        \cmidrule(lr){2-2} \cmidrule(lr){3-3} \cmidrule(lr){4-4} \cmidrule(lr){5-5} \cmidrule(lr){6-6} \cmidrule(lr){7-7} \cmidrule(lr){8-8}
        Methods & ROUGE-L $\uparrow$ & ROUGE-L $\uparrow$ & ROUGE-L $\uparrow$ & Overall Score $\uparrow$ & CIDEr $\uparrow$ & CIDEr $\uparrow$ & CIDEr $\uparrow$ \\
        \midrule
        \multicolumn{8}{c}{LLaVA-OneVision-7B} \\
        \midrule
        Full Tokens & 39.16 & 22.15 & 31.74 & 57.29 & 79.39 & 137.97 & 11.24 \\
        \midrule
        FastV ($K=3,R=0.5$) & 37.41 & 21.16 & 24.78 & 43.22 & 77.38 & 125.01 & 9.59 \\
        FastV ($K=2,R=0.5$) & 36.19 & 20.77 & 23.99 & 43.04 & 75.37 & 120.8 & 9.31 \\
        \midrule
        VTW ($K=20$) & 30.13 & 19.59 & 29.17 & 52.68 & 39.28 & 76.23 & 7.03  \\
        VTW ($K=14$) & 30.47 & 16.17 & 25.35 & 41.47 & 16.80 & 41.43 & 3.03 \\
        \midrule
        PLPHP ($r=0.5$) & \underline{39.72} (+0.56) & \textbf{22.10} & \textbf{31.88} (+0.14)    & \textbf{57.46} (+0.17)    & \textbf{78.93} & \textbf{137.90} & \textbf{10.43} \\
        PLPHP ($r=0.4$) & \textbf{39.81}(+0.65)    & \underline{22.06} & \underline{31.82} (+0.08) & \underline{57.41} (+0.12) & \underline{78.55} & \underline{137.64} & \underline{9.89} \\
        \midrule
        \multicolumn{8}{c}{LLaVA-OneVision-0.5B} \\
        \midrule
        Full Tokens & 36.37 & 17.12 & 29.76 & 54.01 & 75.39 & 129.87 & 10.45 \\
        \midrule
        FastV ($K=3,R=0.5$) & 23.06 & 12.87 & 24.97 & 39.03 & 64.22 & 97.74 & 8.25 \\
        FastV ($K=2,R=0.5$) & 21.81 & 11.18 & 24.51 & 34.15 & 61.97 & 98.73 & 7.91 \\
        \midrule
        VTW ($K=17$) & 24.43 & \textbf{16.91} & 26.96 & 41.16 & 12.79 & 14.54 & 2.38 \\
        VTW ($K=12$) & 24.74 & 16.51 & 24.35 & 46.60 & 7.35 & 9.80 & 1.25 \\
        \midrule
        PLPHP ($r=0.5$) & \textbf{36.35} & 16.81 & \underline{29.88} (+0.12) & \textbf{54.01} & \textbf{72.34} & \textbf{126.72} & \textbf{9.31} \\
        PLPHP ($r=0.4$) & \underline{36.19} & \underline{16.82} & \textbf{30.03} (+0.27) & \underline{53.91} & \underline{71.04} & \underline{123.75} & \underline{8.35} \\
        \bottomrule
    \end{tabular}}
    \label{tab:main-results}
\end{table*}

\begin{figure*}[h!]
	\centering
	\subfloat[Spot-the-Diff]{
		\includegraphics[width=0.22\textwidth]{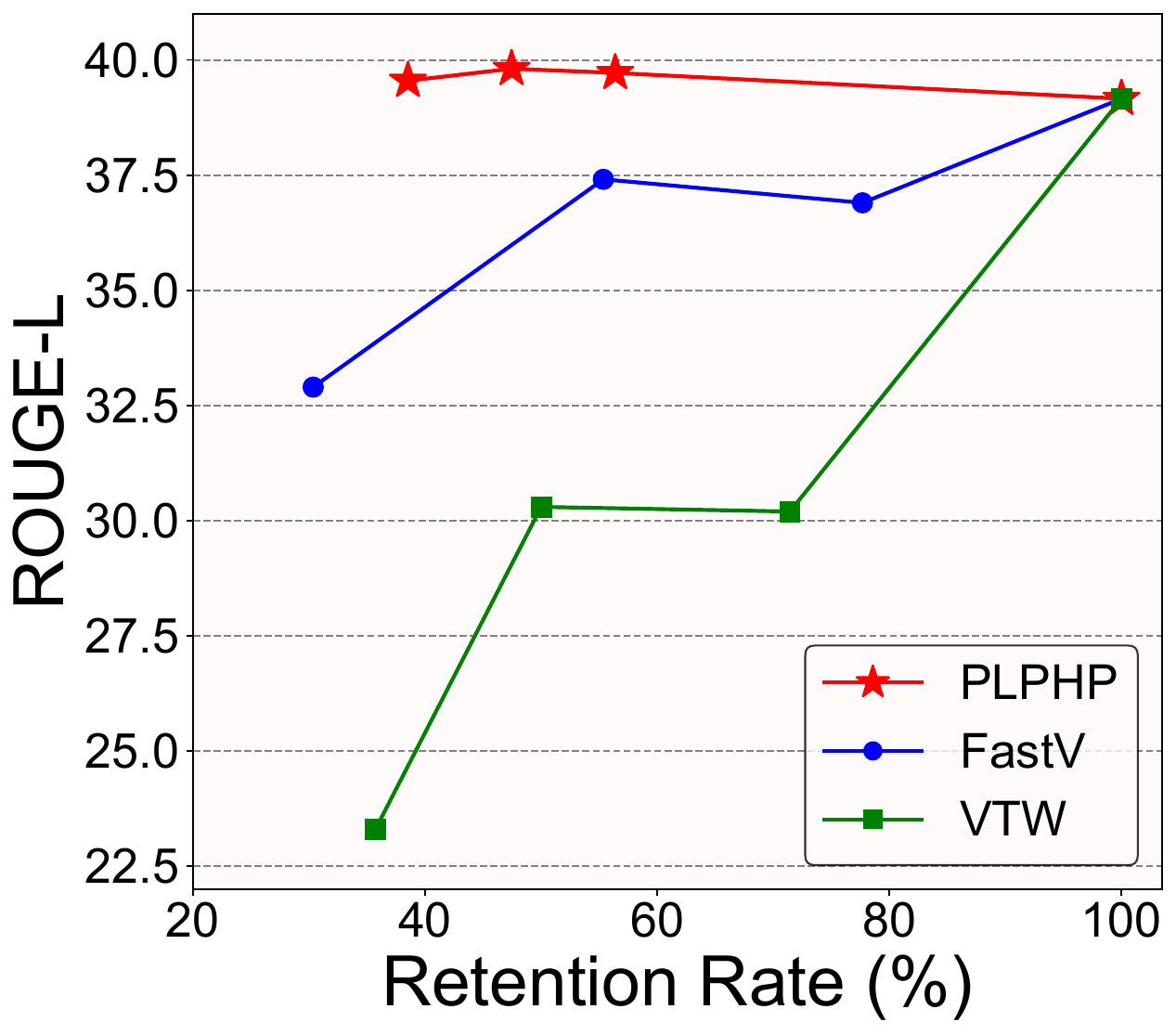}}
	\subfloat[Image-Edit]{
		\includegraphics[width=0.22\textwidth]{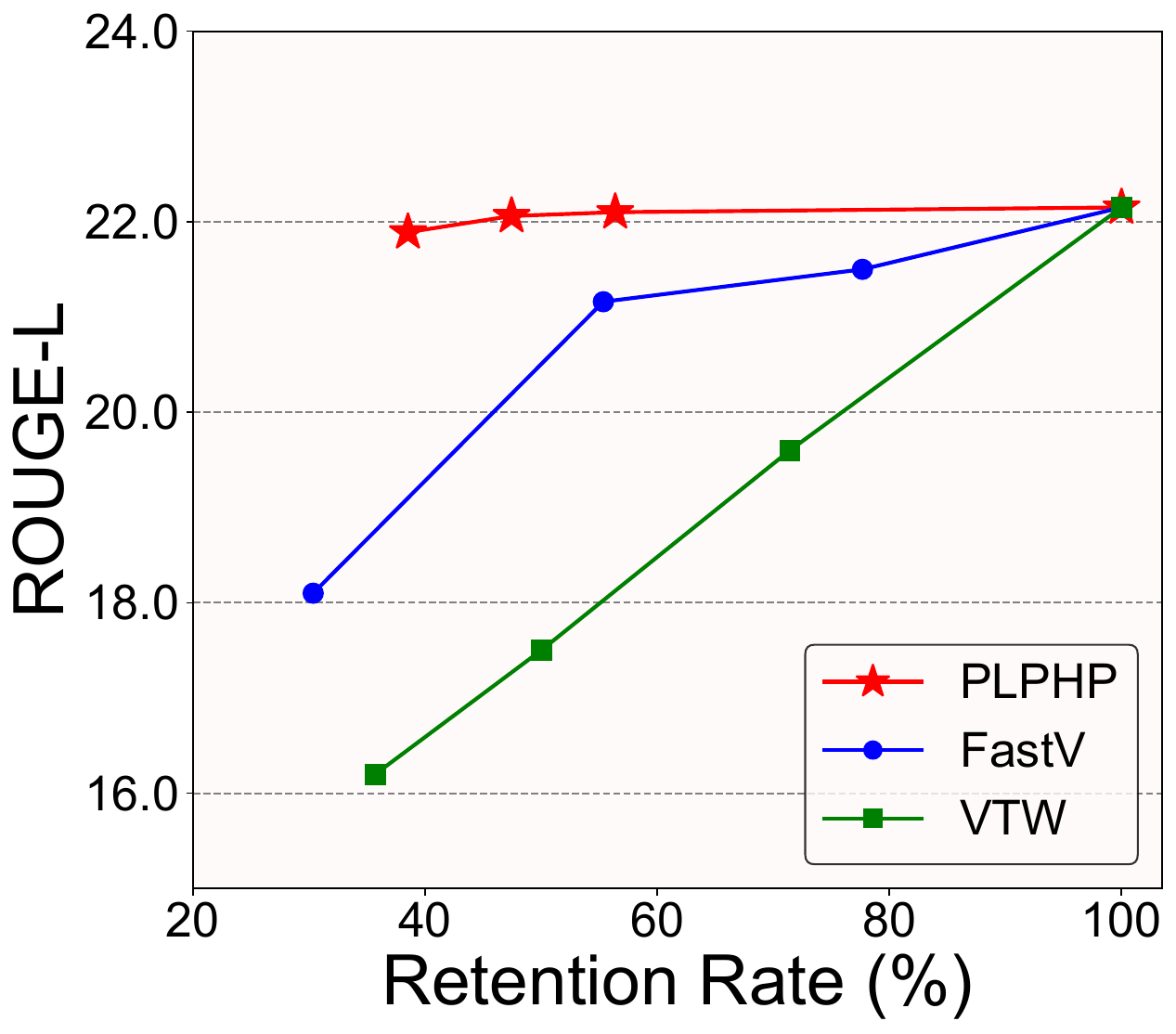}}
	\subfloat[Visual-Story-Telling]{
		\includegraphics[width=0.22\textwidth]{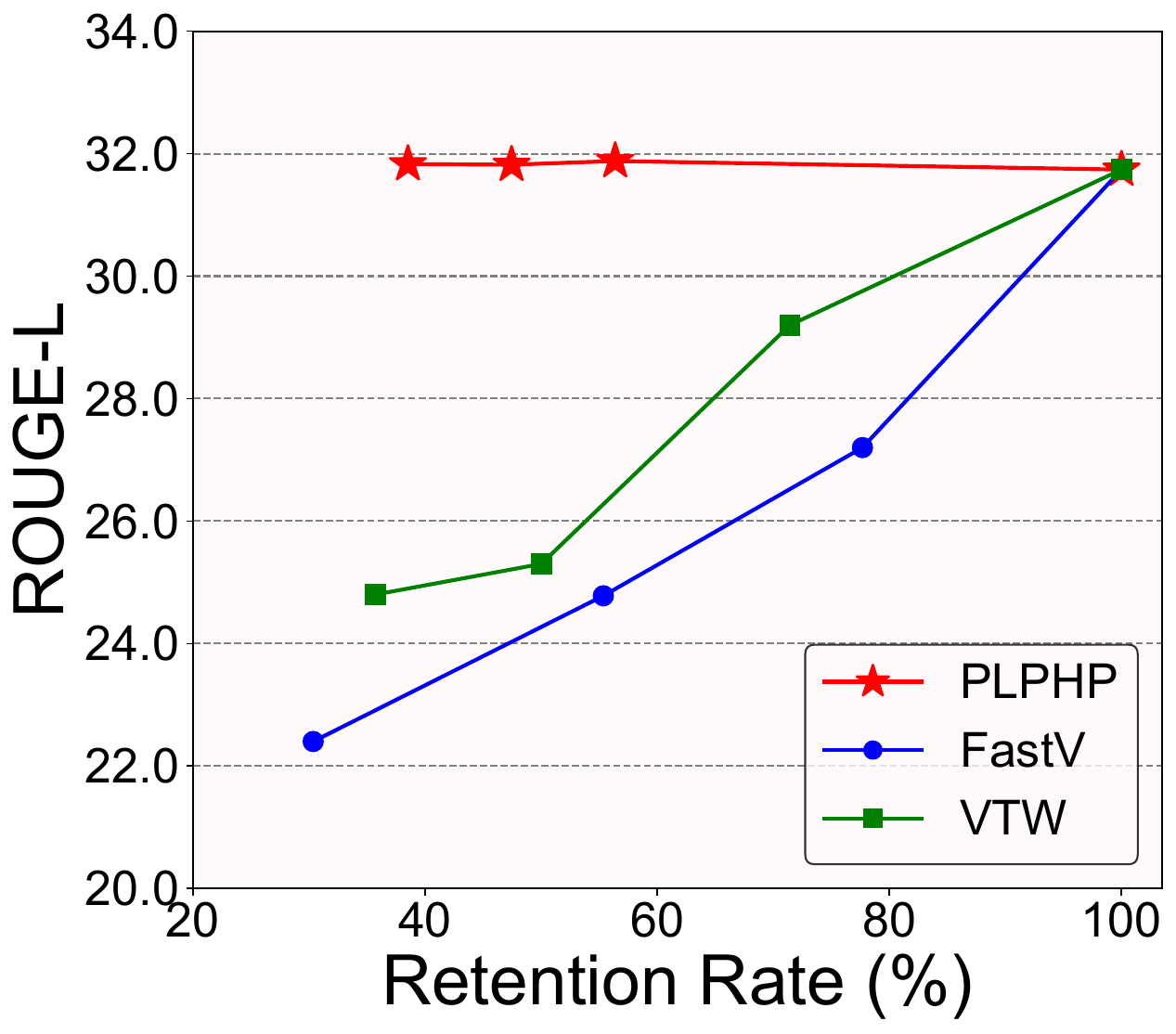}}
	\subfloat[Multi-View]{
		\includegraphics[width=0.22\textwidth]{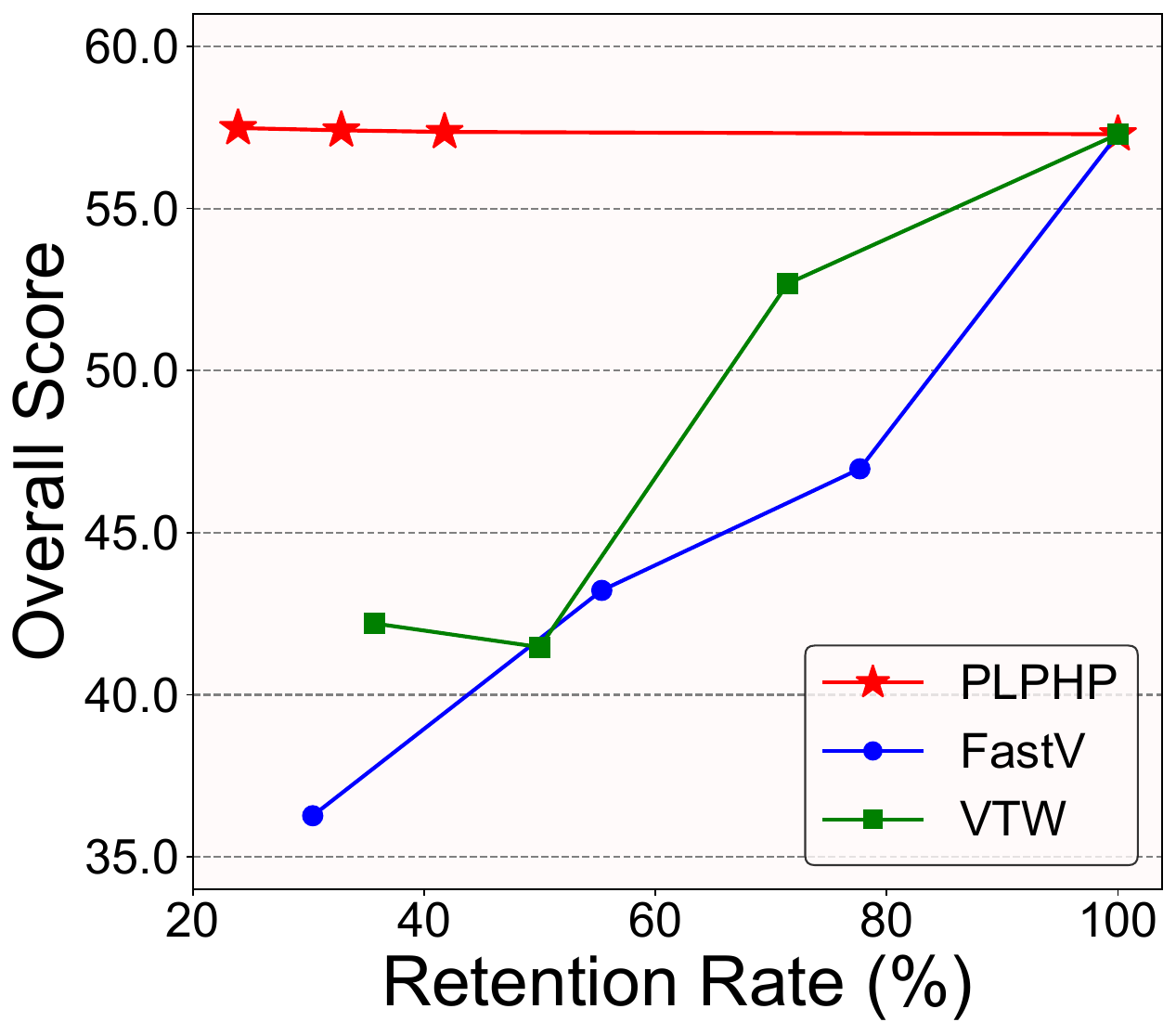}}
        \\ \quad \\ \quad \\
	\subfloat[Flickr30k]{
		\includegraphics[width=0.22\textwidth]{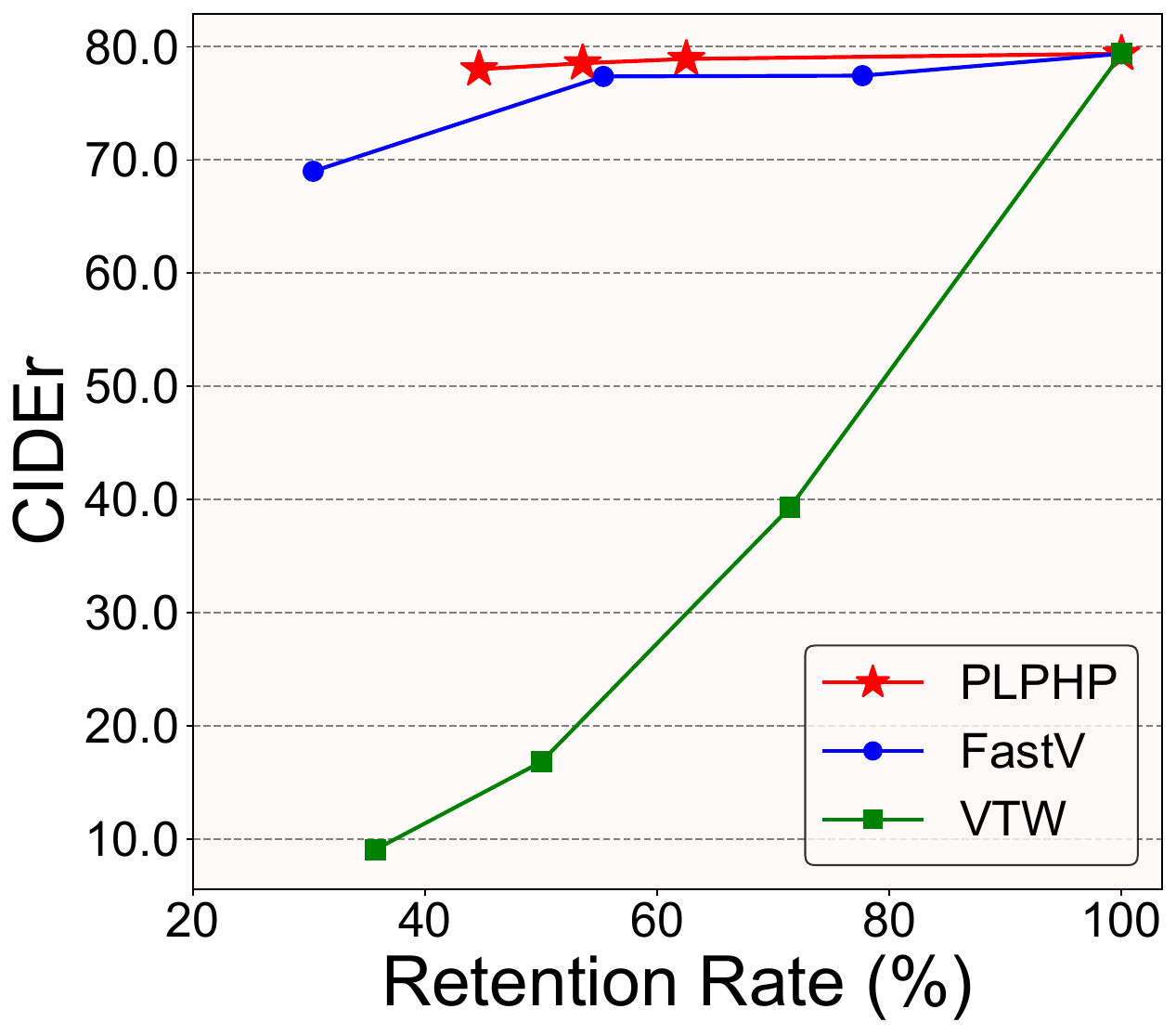}}
	\subfloat[COCO 2017 Caption]{
		\includegraphics[width=0.22\textwidth]{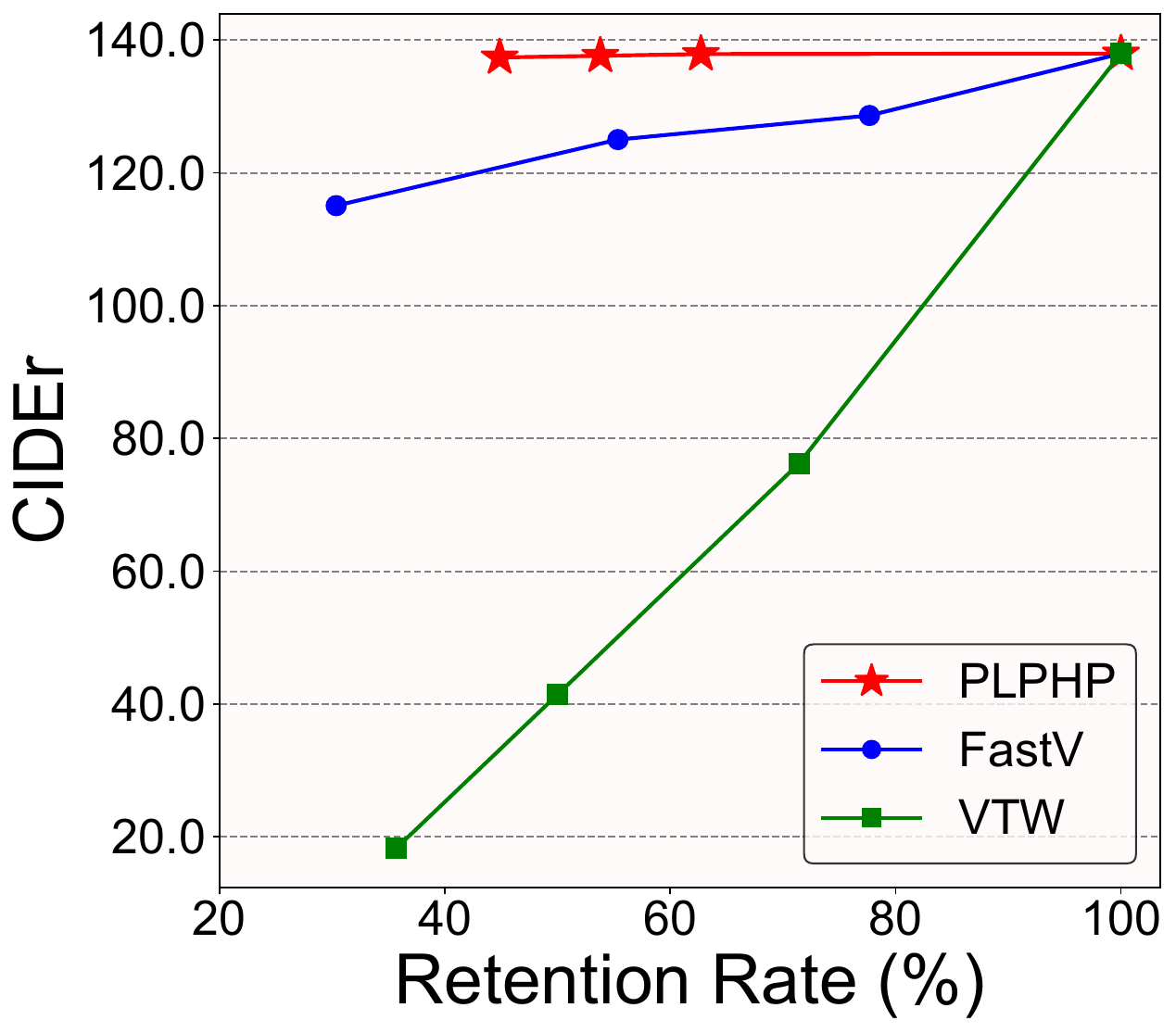}}
	\subfloat[DetailCaps4870]{
		\includegraphics[width=0.22\textwidth]{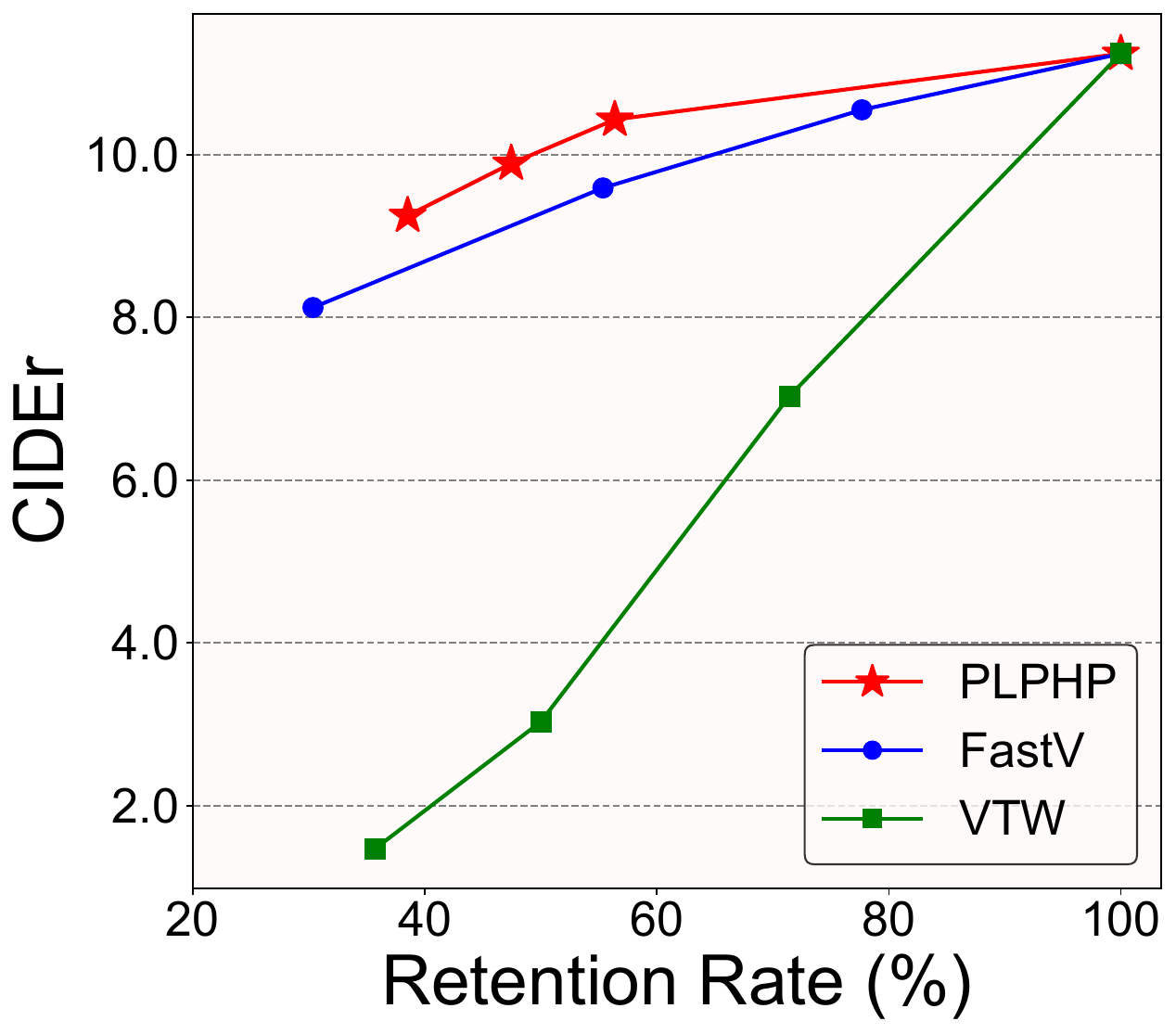}}
	\caption{\textbf{Visualization of vision token retention rates and performance across seven different benchmarks}. A point on each polyline represents a certain hyperparameter setting. We record the vision token retention rate and performance of the method under the corresponding setting. For VTW, we evaluated cases with $K=10, 14$ and $20$. For FastV, we assessed the cases of $(K, R)=(2,0.75), (3,0.5)$ and $(3,0.25)$. As for PLPHP, we examined the situations where $(r, \Delta r)=(0.3,0.3), (0.4,0.3)$ and $(0.5,0.3)$.}
        \label{fig:main}
\end{figure*}

\begin{figure*}[h!]
	\centering
	\subfloat[DetailCaps4870]{
		  \includegraphics[width=0.22\textwidth]{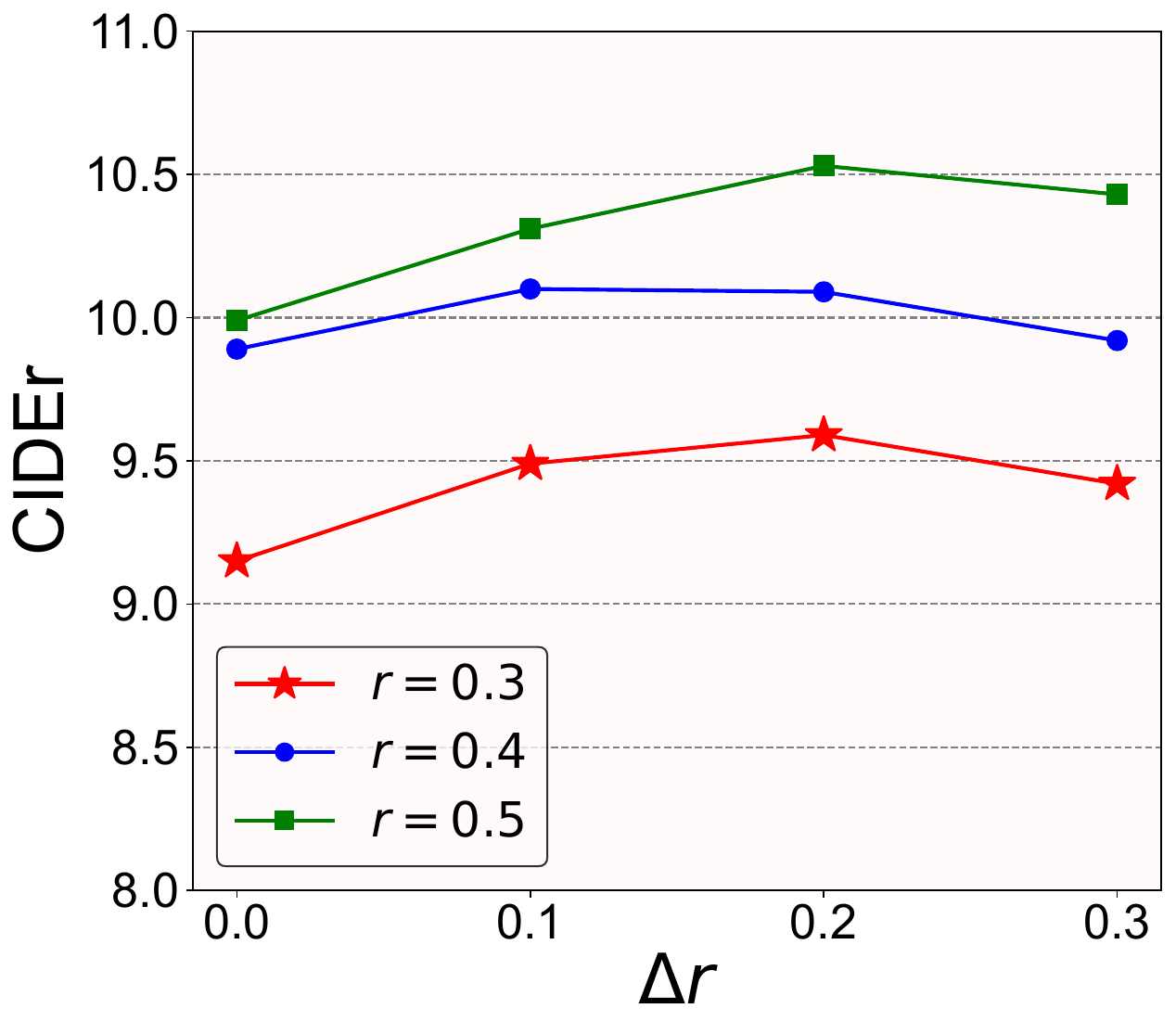}\label{fig:ab-dc}
        }
	\subfloat[Spot-the-Diff]{
		  \includegraphics[width=0.22\textwidth]{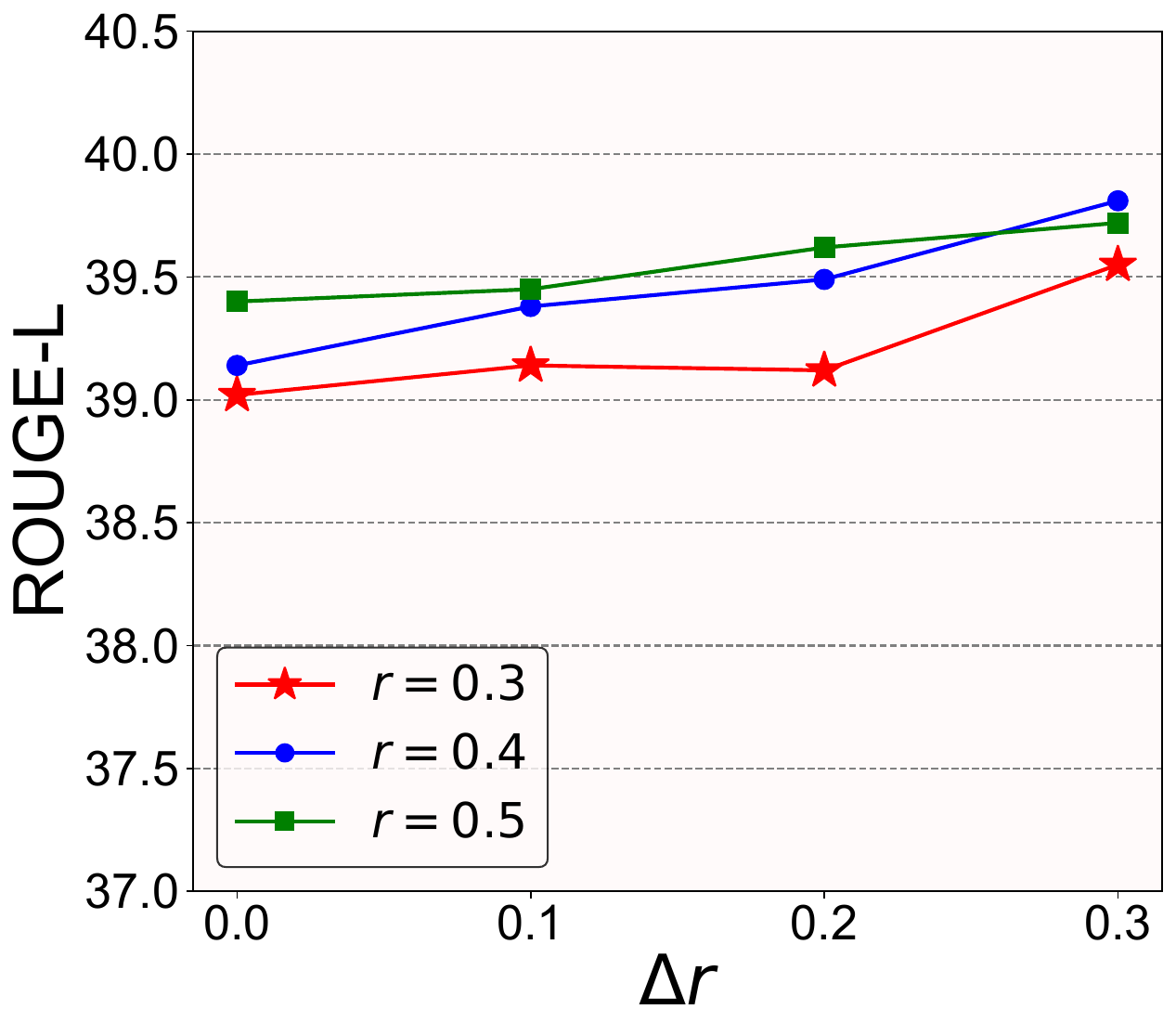}\label{fig:ab-SD}
        }
	\subfloat[Image-Edit]{
		  \includegraphics[width=0.22\textwidth]{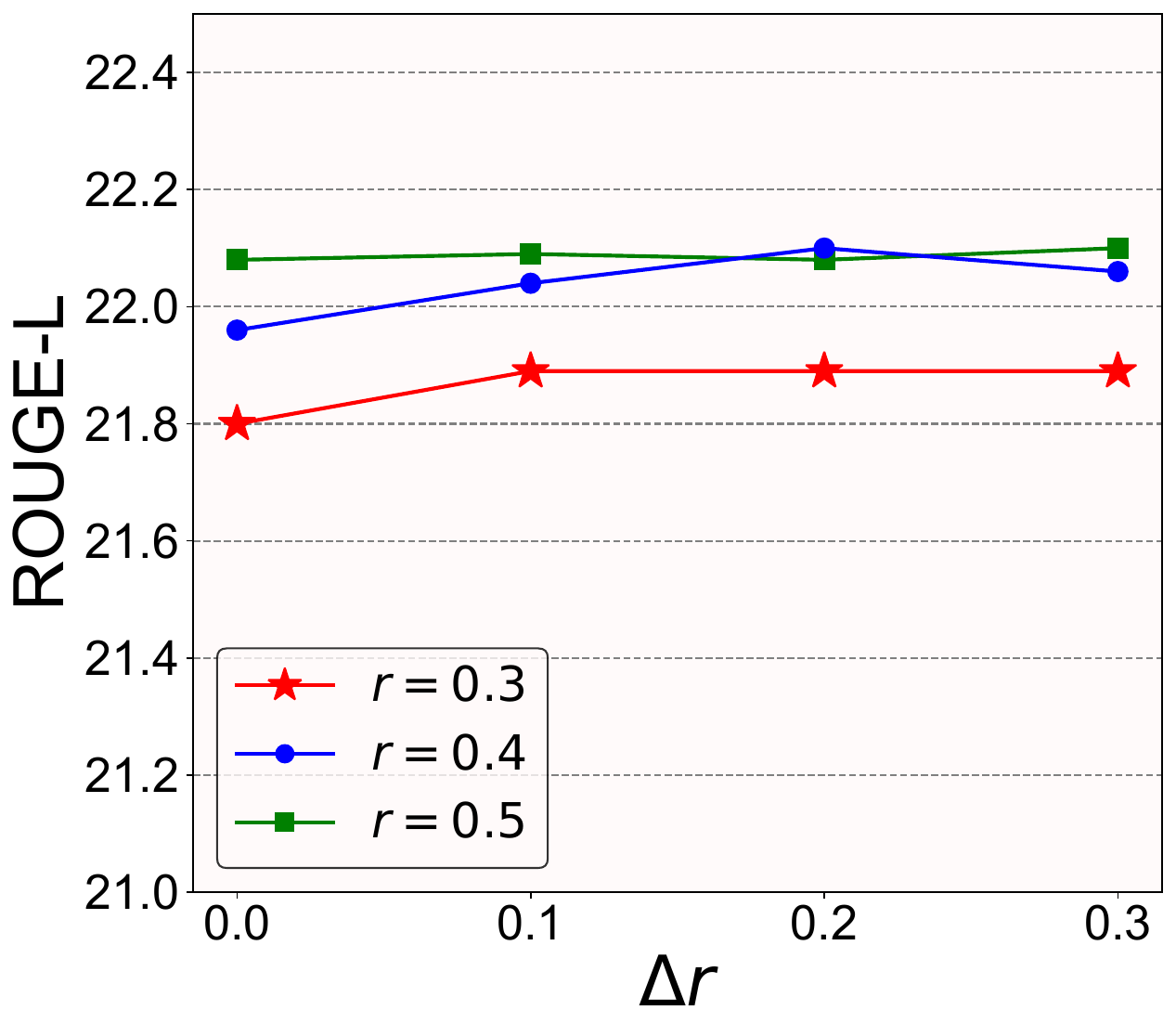}\label{fig:ab-IE}
        }
	\subfloat[Visual-Story-Telling]{
		  \includegraphics[width=0.22\textwidth]{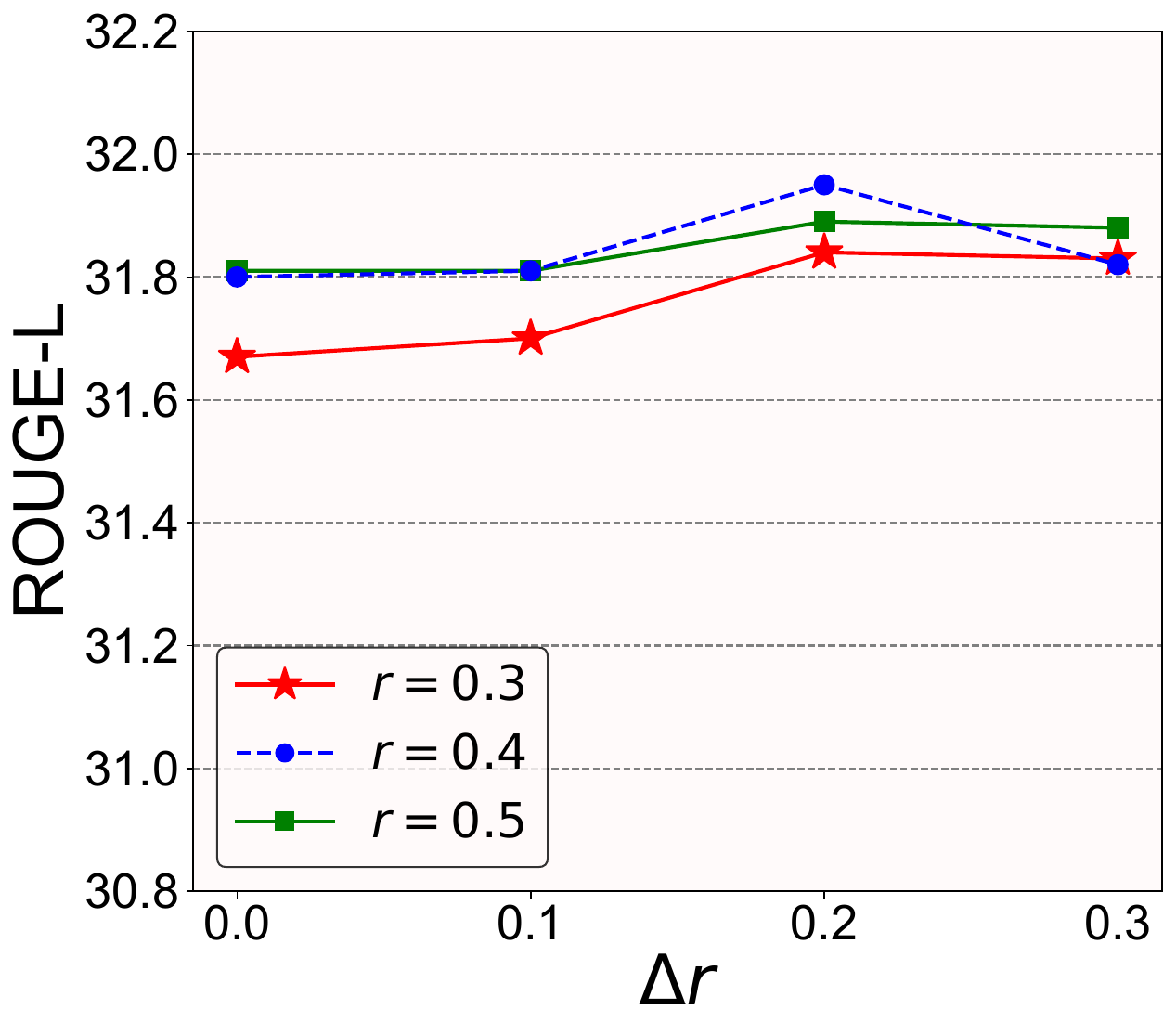}\label{fig:ab-VST}
        }
	\caption{\textbf{Ablation studies on $r$ and $\Delta r$.} Each polyline in the figure corresponds to a specific value of $r$, with different points on a single line representing various values of $\Delta r$ and their corresponding performance metrics.}
		\label{fig:ablation-r-dr}
\end{figure*}

\subsection{Experimental Setting}

\noindent \textbf{Benchmarks.} 
In terms of multi-image benchmarks, we select four subsets from LLaVA-NeXT-Interleave-Bench \cite{li2024llava-interleave}: Spot-the-Diff (SD), Image-Edit (IE), Visual-Story-Telling (VST), and Multi-View (MV). We also select three single-image benchmarks: Flickr30k \cite{plummer2015flickr30k}, COCO 2017 Caption\cite{lin2014microsoft}, and DetailCaps4870 \cite{dong2024benchmarking}.

\noindent \textbf{Metrics.} Open-ended VQA tasks are evaluated using the ROUGE-L \cite{lin2004rouge} (R) metric. CIDEr \cite{vedantam2015cider} (C) and METEOR \cite{banerjee2005meteor} (M) are employed to assess image captioning tasks. Overall Score is used to evaluate the performance on Multi-View benchmark. Regarding efficiency analysis, we utilize Vision Token Retention Rate (RR), KV Cache Size (KV), and Decoding Latency as our metrics for evaluation.

\noindent \textbf{Baselines.} 
We choose FastV and VTW as our baselines. FastV discards image tokens with low attention scores in the shallow layers, while VTW retains all image tokens in the shallow layers and discards them in the deeper layers.

\noindent \textbf{Implementation Details.} 
We implement PLPHP and all baselines on an NVIDIA A100 (80GB) GPU. All methods are evaluated using LMMs-Eval \cite{lmms_eval2024,zhang2024lmmsevalrealitycheckevaluation}. More discussions regarding our benchmark selection, baseline configuration, and implementation details can be found in Appendix \ref{asec:eval-setting}.

Unless otherwise specified, the experimental results we report are based on LLaVA-OneVision-7B, and the default hyperparameter setting of PLPHP is $(r, \Delta r, \alpha, \beta) = (0.4, 0.3, 0.25, 0.1)$. The bolded text in the tables indicates the \textbf{best} performance under the corresponding metric, while the underlined text denotes the \underline{second best}.

\begin{table*}[h!]
    \centering
    \caption{\textbf{Ablation studies on $\alpha$ and $\beta$.}}
    \vspace{-0.3cm}
    \renewcommand{\arraystretch}{1}
    \resizebox{\textwidth}{!}{
    \begin{tabular}{lcccccc}
        \toprule
        & Spot-the-Diff & Image-Edit & Visual-Story-Telling & DetailCaps &  &  \\
        \cmidrule(lr){2-2} \cmidrule(lr){3-3} \cmidrule(lr){4-4} \cmidrule(lr){5-5} \cmidrule(lr){6-7} 
        Methods & ROUGE-L $\uparrow$ & ROUGE-L $\uparrow$ & ROUGE-L $\uparrow$ & CIDEr $\uparrow$ & Avg. Retention Rate (\%) $\downarrow$ & Avg. KV Cache Size (\%) $\downarrow$ \\
        \midrule
        $\alpha=0.25,\beta=0.05$ & \underline{39.74} & \textbf{22.10} & \underline{31.82} & \textbf{10.66} & 50.6\% & 53.2\% \\
        $\alpha=0.2,\beta=0.1$ & 39.15 & \textbf{22.10} & \textbf{31.87} & \underline{10.16} & 44.0\% & 50.4\% \\
        $\alpha=0.25,\beta=0.1$ & \textbf{39.81} & 22.06 & \underline{31.82} & 9.89 & 41.6\% & 47.7\% \\
        $\alpha=0.3,\beta=0.1$ & 39.35 & 22.02 & 31.81 & 9.63 & \underline{39.6\%}  & \underline{45.1\%} \\
        $\alpha=0.25,\beta=0.15$ & 39.51 & \underline{22.07} & 31.80 & 9.55 & \textbf{35.8}\% & \textbf{42.6\%} \\
        \bottomrule
    \end{tabular}
    }
    \vspace{-0.5cm}
    \label{tab:ablation-alpha-beta}
\end{table*}

\subsection{Main Results}

We first conduct experiments with our method based on LLaVA-OneVision across different benchmarks. The main results are shown in Table \ref{tab:main-results}. From the table, we can observe that:
\begin{itemize}[leftmargin=*]
\setlength{\topmargin}{0pt}
\setlength{\itemsep}{0em}
    \item \textbf{PLPHP significantly outperforms both baselines across different benchmarks.} For the LLaVA-OneVision-7B model, the average performance of PLPHP under default hyperparameter settings surpasses FastV by 11.4\% and VTW by 48.4\%. Compared to the uncompressed model, the average performance degradation brought by PLPHP is merely 0.46\%. We attribute this performance enhancement to the granularity and adaptability of PLPHP. In contrast to FastV and VTW, which discard a fixed set of vision tokens from \textit{all} pruned attention heads, the dynamic nature of PLPHP offers a distinct performance advantage.
    \item \textbf{Model with PLPHP outperforms uncompressed model on various multi-image tasks.} Notably, the average performance of PLPHP surpasses that of the uncompressed model by 0.51\% across multiple multi-image task benchmarks on LLaVA-OneVision-7B through appropriate pruning. The improvement on multi-image benchmarks could be attributed to the increased redundancy in visual information inherent in multi-image tasks, which could potentially be detrimental to model inference. This redundancy is effectively eliminated by PLPHP, thereby enhancing both the efficiency and performance. 
    \item \textbf{The performance of PLPHP remains relatively stable under different retention rates.} The carefully designed pruning dynamics in PLPHP allow it to prioritize the removal of the most redundant tokens, thereby ensuring that performance is less affected by the pruning rate. On the other hand, VTW is highly sensitive to the selection of $K$. It discards \textit{all} vision tokens at a specific layer, thus once the model exhibits significant \textit{Vision Token Re-attention} after this layer, it is likely to severely impact the performance, which could be the cause of its high sensitivity to the hyperparameter and substantial performance decline in image captioning tasks.
\end{itemize}

To provide a more intuitive analysis of how each method performs under varying pruning rates, we evaluated their performance across different vision token retention rates and visualized the results in Figure \ref{fig:main}. It can be observed that PLPHP consistently outperforms the baseline at the same pruning rate and maintains nearly no performance degradation within a certain pruning rate range, indicating that we can achieve better performance while discarding more vision tokens, which directly leads to a higher computational efficiency.

These performance boosts highlight the superiority of our method, which dynamically adjusts the pruning rate based on the attention allocated to image tokens in different layers and independently preserve different contextual information for different attention heads.

\begin{table}[h]
    \centering
    \caption{\textbf{Performance of PLPHP on various models.} Bolded text indicates that PLPHP surpasses the uncompressed model.}
    \vspace{-0.3cm}
    \renewcommand{\arraystretch}{1.1}
    \resizebox{0.5\textwidth}{!}{
    \begin{tabular}{lcccccccc}
        \toprule
        & SD
        & IE
        & VST
        & MV 
        & Flickr30k
        & COCO
        & 
        & \\
        \cmidrule(lr){2-2} \cmidrule(lr){3-3} \cmidrule(lr){4-4} \cmidrule(lr){5-5} \cmidrule(lr){6-6} \cmidrule(lr){7-7} \cmidrule(lr){8-9} 
        Methods
        & R $\uparrow$ 
        & R $\uparrow$ 
        & R $\uparrow$
        & R $\uparrow$
        & C $\uparrow$ 
        & C $\uparrow$
        & RR (\%) $\downarrow$
        & KV (\%) $\downarrow$\\
        \midrule
        Qwen2-VL  & 27.56 & 21.21 & 24.92 & 12.78 & 77.24 & 96.18 & 100\% & 100\% \\
         w/ PLPHP & \textbf{27.78} & \textbf{21.40} & \textbf{25.02} & \textbf{12.96} & \textbf{78.02} & \textbf{98.67} & \textbf{35.8\%} & \textbf{41.9\%} \\
        \midrule
        IDEFICS2 & 18.98 & 14.90 & 23.91 & 13.84 & 51.73 & 72.12 & 100\% & 100\% \\
        w/ PLPHP & 18.55 & 14.89 & \textbf{23.93} & \textbf{13.96} & 51.68 & \textbf{72.60} & \textbf{36.1\%} & \textbf{51.3\%} \\
        \midrule
        Mantis   & 16.30 &  9.56 & 13.27 & 11.02 & 70.41 & 91.41 & 100\% & 100\% \\
        w/ PLPHP & \textbf{16.41} & \textbf{9.81} & \textbf{13.41} & \textbf{11.14} & 69.90 & 90.61 & \textbf{29.1\%} & \textbf{33.7\%} \\
        \bottomrule
    \end{tabular}
    }
    \vspace{-0.5cm}
    \label{tab:performance-on-different-models}
\end{table}

\subsection{Generality of PLPHP on Various LVLMs}

To further demonstrate the generality of PLPHP on various model architectures, we implement PLPHP on common LVLMs with different LLM backbones, and directly compared them with uncompressed models to highlight our effectiveness, with results shown in Table \ref{tab:performance-on-different-models}. Since IDEFICS2 and Mantis are unable to follow instructions in DetailCaps4870, we evaluate PLPHP on the other six benchmarks. \textbf{Remarkably, Qwen2-VL equipped with PLPHP surpasses the uncompressed model across all benchmarks}, achieving an average improvement rate of 1.5\%, while saving an average of 58.1\% KV Cache storage space. For the other two models, our method also achieves an average of 57\% KV Cache compression while surpassing the original models across multiple benchmarks.

\begin{figure}[h]
	\centering
        \vspace{-0.2cm}
        \subfloat[Decoding Latency]{
		  \includegraphics[width=0.23\textwidth]{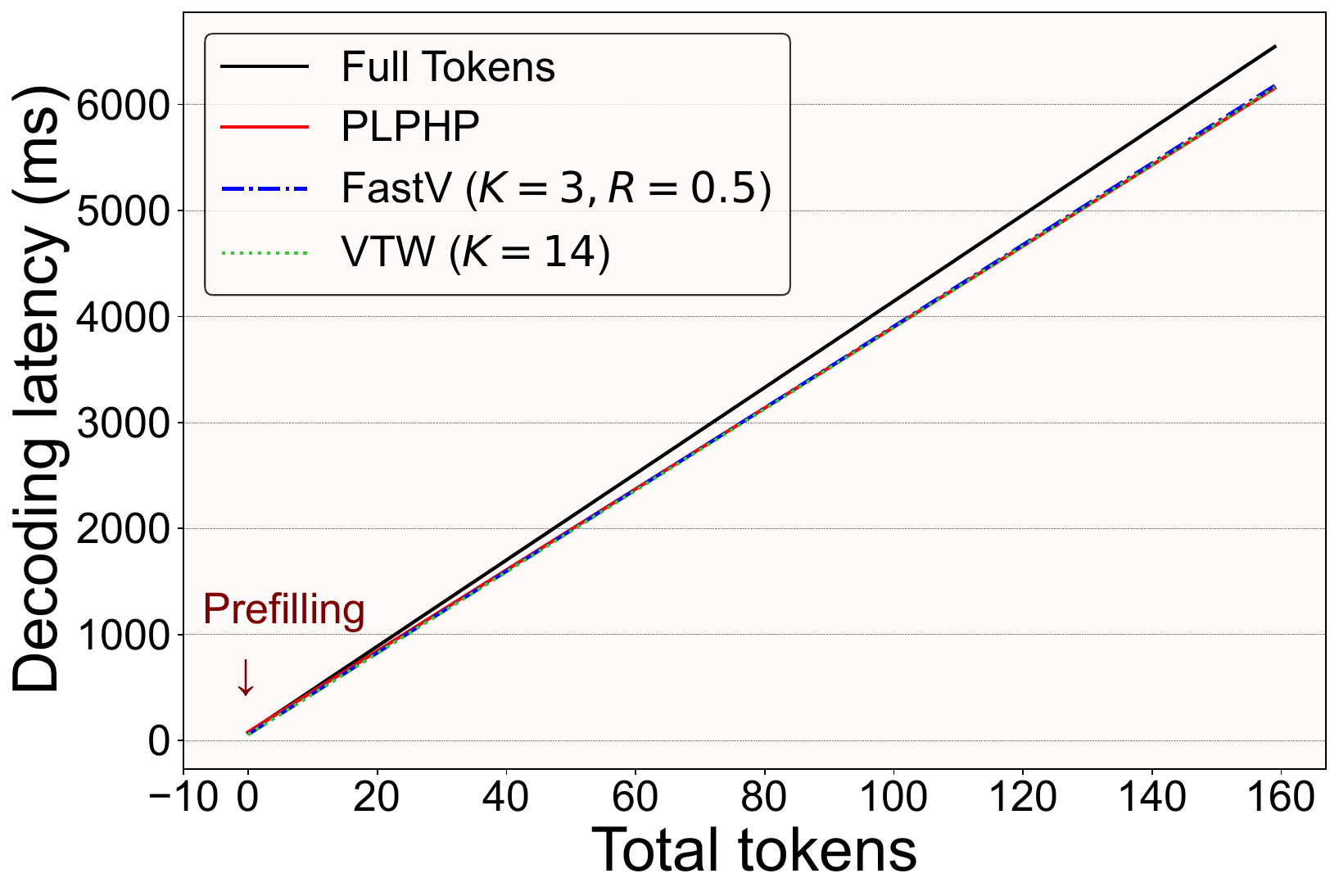}\label{fig:efficiency-latency}
        }
        \subfloat[KV Cache Size]{
		  \includegraphics[width=0.23\textwidth]{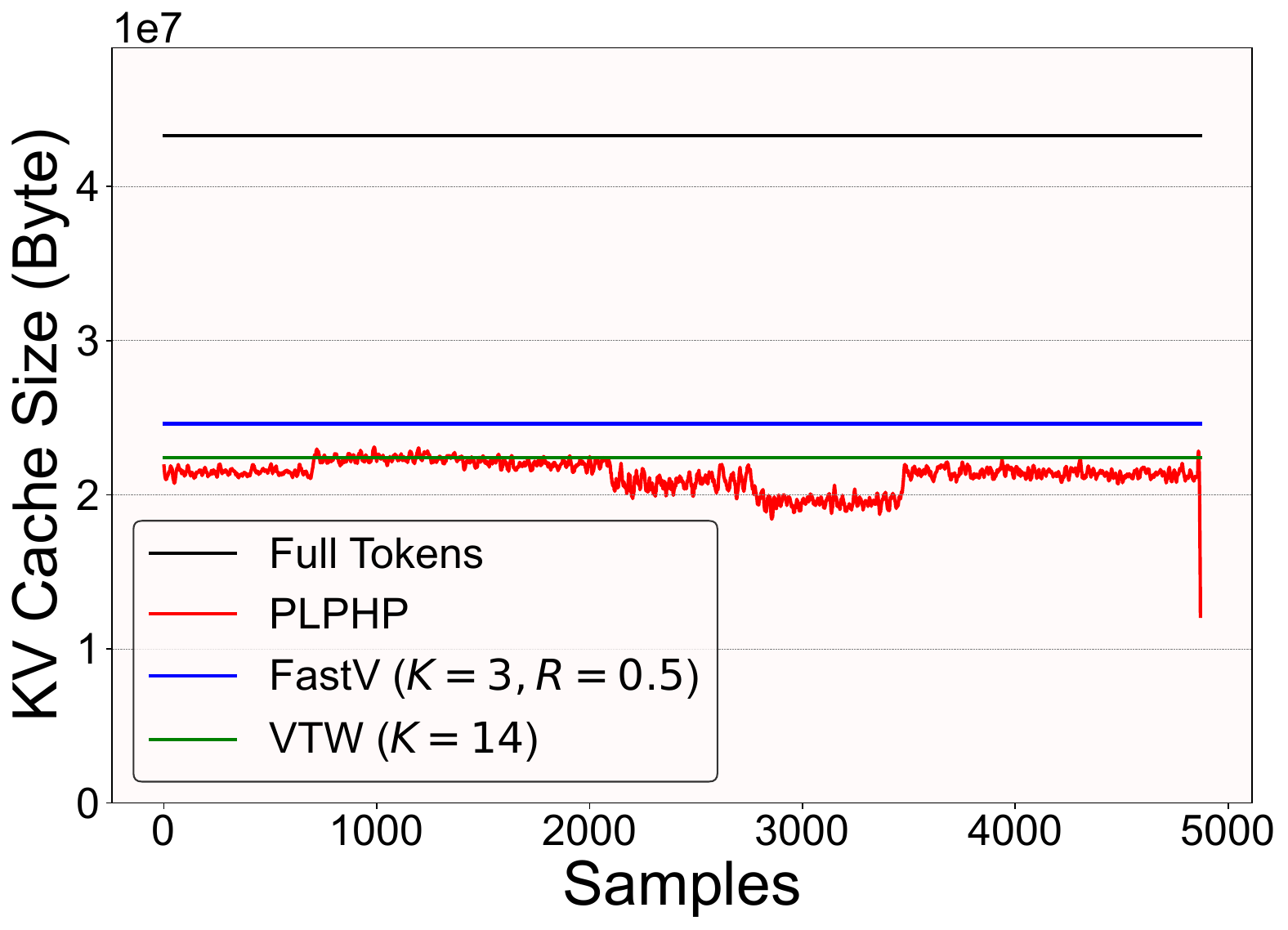}\label{fig:efficiency-kv}
        }
        \vspace{-0.1cm}
	\caption{\textbf{The decoding latency and KV Cache size results.} Both baselines maintain constant KV Cache sizes due to unchanging pruning rates, while PLPHP adaptively assigns retention rates, producing a fluctuating curve with a smaller mean.}
        \vspace{-0.4cm}
		\label{fig:efficiency-latency-kv}
\end{figure}

\begin{table}[h]
    \centering
    \caption{\textbf{Performance and efficiency comparison among different methods.}}
    \vspace{-0.3cm}
    \renewcommand{\arraystretch}{1}
    \resizebox{0.5\textwidth}{!}{
    \begin{tabular}{lccccc}
        \toprule
        & \multicolumn{5}{c}{DetailCaps4870} \\
        \cmidrule(lr){2-6}
        Methods & C $\uparrow$ & M $\uparrow$ & R $\uparrow$ & Time (h) $\downarrow$ & RR (\%) $\downarrow$ \\
        \midrule
        Full Tokens & 11.24 & 20.13 & 30.01 & 5.63 & 100\%\\
        \midrule
        FastV ($K=3, R=0.5$) & \underline{9.59} & \underline{18.55} & \underline{28.72} & \underline{5.23} & 55.4\%\\
        VTW ($K=14$) & 3.03 & 15.05 & 24.38 & \textbf{5.22} & \underline{50.0\%} \\
        PLPHP & \textbf{9.89} & \textbf{19.33} & \textbf{29.19} & \underline{5.23} & \textbf{47.5\%}\\
        \bottomrule
    \end{tabular}
    }
    \vspace{-0.4cm}
    \label{tab:efficiency-on-detailcaps}
\end{table}

\subsection{Efficiency Analysis}

To analyze the efficiency of PLPHP, we conduct experiments on DetailCaps4870 since it includes long generation contents. We can observe from Figure \ref{fig:efficiency-latency} that PLPHP achieves a comparable total decoding latency to both baselines. The latency introduced by the unpruned Prefilling Stage is minimal (less than 0.5 tokens of delay). Figure \ref{fig:efficiency-kv} shows that PLPHP maintains a lower KV cache size during the evaluation process compared to all baselines, leading to a shorter decoding latency. Table \ref{tab:efficiency-on-detailcaps} shows that PLPHP attains performance closest to the uncompressed model. The nearly consistent evaluation time also indicates that the additional computation during the Prefilling Stage gradually becomes negligible as generation progresses.

\begin{table}[h]
    \centering
    \caption{\textbf{Decoding Latency and KV Cache Size of PLPHP under different retention rates.}}
    \vspace{-0.3cm}
    \renewcommand{\arraystretch}{1.1}
    \resizebox{0.5\textwidth}{!}{
    \begin{tabular}{lcc}
        \toprule
        Methods  & Decoding Latency (ms/token) $\downarrow$ & KV Cache Size (\%) $\downarrow$ \\
        \midrule
        Full Tokens &  49.10 &  100\% \\
        \midrule
        PLPHP ($r=0.5$)  & 41.26 & 54.9\% \\
        PLPHP ($r=0.4$) & \underline{40.20} & \underline{46.2\%} \\
        PLPHP ($r=0.3$) & \textbf{39.19} & \textbf{37.6\%} \\
        \bottomrule
    \end{tabular}
    }
    \vspace{-0.5cm}
    \label{tab:efficiency-analysis}
\end{table}

\subsection{Ablation Study}

To explore the impact of $r$ and $\Delta r$, we conduct ablation experiments on four benchmarks, with the results illustrated in Figure \ref{fig:ablation-r-dr}. It can be observed that setting $\Delta r > 0$ consistently outperforms the cases where $\Delta r=0$, indicating that adaptive pruning rates are superior to a fixed pruning rate. This finding demonstrates that our proposed \textbf{layer-level pruning rate allocation has a positive impact on model performance}.

Since $r$ is the most direct parameter reflecting the average pruning rate, we test the impact of $r$ on efficiency, with the results presented in Table \ref{tab:efficiency-analysis}. PLPHP achieves an 18.1\% decoding speedup and a 53.8\% KV Cache compression under the default settings where $r=0.4$, and further reaches a 20.2\% acceleration and a 62.4\% compression at a lower retention rate, enhancing the computational efficiency of LVLM decoding remarkably.

$\alpha$ and $\beta$ also indirectly influence pruning rates, thus we also conduct ablation studies with the results shown in Table \ref{tab:ablation-alpha-beta}. Intuitively, increasing $\alpha$ and $\beta$ elevates the criteria for vision-attentive layers and vision-balanced layers more stringent, leading to higher pruning rates at the cost of performance loss. Conversely, decreasing them relaxes the criteria, enhancing the performance but at greater computational expense.

\section{Conclusion}\label{sec::conclusion}
In this work, we introduce PLPHP, a two-level pruning method designed to improve the efficiency of LVLMs with Layer-Level Retention Rate Allocation and Head-Level Vision Token Pruning. Comprehensive experiments demonstrate that PLPHP outperforms existing pruning methods, achieving a 18\% decoding acceleration, over 50\% KV Cache compression and only 0.46\% performance degradation, with improvements on multi-image tasks. We believe our work contributes to efficient LVLMs, further promotes their applications, and improves the user experience.

\clearpage

\bibliography{bibliography}

\appendix
\cleardoublepage
\section{Appendix}

\subsection{Details of Evaluation Settings}\label{asec:eval-setting}

\subsubsection{Benchmarks}\label{asec:eval-setting-bench}

Since PLPHP maintains the computational integrity of the LVLMs' Prefilling Stage, its efficiency advantage is primarily reflected in the low decoding latency during the subsequent Decoding Stage. Therefore, we mainly choose benchmarks composed of open-ended VQA and image captioning tasks. The benchmarks we select encompasses both multi-image task benchmarks and single-image task benchmarks.

$\bullet$ \textbf{Multi-Image benchmarks}: The LLaVA-Interleave Bench is a comprehensive benchmark dataset designed to evaluate the performance of LVLMs in multi-image scenarios. It consists of 13 challenging tasks with a total of 17,000 instances. We curated four subsets consisting of open-ended VQA tasks from LLaVA-NeXT-Interleave-Bench: Spot-the-Diff, Image-Edit, Visual-Story-Telling, and Multi-View.

$\bullet$ \textbf{Single-Image benchmarks}: The Flickr30k dataset is a widely used benchmark in the field of image captioning and visual understanding. It consists of 31,783 images collected from the Flickr platform, each paired with five human-annotated captions. The COCO2017 Caption subset contains more than 45,000 images, each annotated with five captions written by human annotators, describing the visual content of the images in detail, including objects, their attributes, and the relationships between them. DetailCaps4870 provides more fine-grained and specific image content descriptions than standard captioning datasets, which is more useful for efficiency analysis. 

\subsubsection{Baselines}\label{asec:eval-setting-baseline}

We select FastV and VTW as our baselines in our experiments. Notably, FastV offers two versions of implementation: one that supports KV cache and one that does not. Since the non-KV-cache implementation introduces substantial computational overhead, we use the version that supports KV cache to ensure a fair comparison. For both of the baselines, we refer to the official open source code \footnote{\url{https://github.com/pkunlp-icler/FastV}} \footnote{\url{https://github.com/lzhxmu/VTW}} and implement them on the models we evaluate.

\subsubsection{Models}\label{asec:eval-setting-impl}

For Qwen2-VL, we set \texttt{max\_pixels} to $1280 \times 28 \times 28$ and \texttt{min\_pixels} to $256 \times 28 \times 28$ according to the official recommendation. The Mantis model that we choose is Mantis-8B-SigLIP-LLaMA3. For LLaVA-OneVision and Mantis, we use the official original versions \footnote{\url{https://huggingface.co/lmms-lab/llava-onevision-qwen2-7b-ov}} \footnote{\url{https://huggingface.co/TIGER-Lab/Mantis-8B-siglip-llama3}}, while using the versions provided by the transformers library \cite{wolf-etal-2020-transformers} for all other models.

\subsection{Case Study}

\begin{figure*}[ht]
	\centering
	\subfloat[]{
		\includegraphics[width=0.48\textwidth]{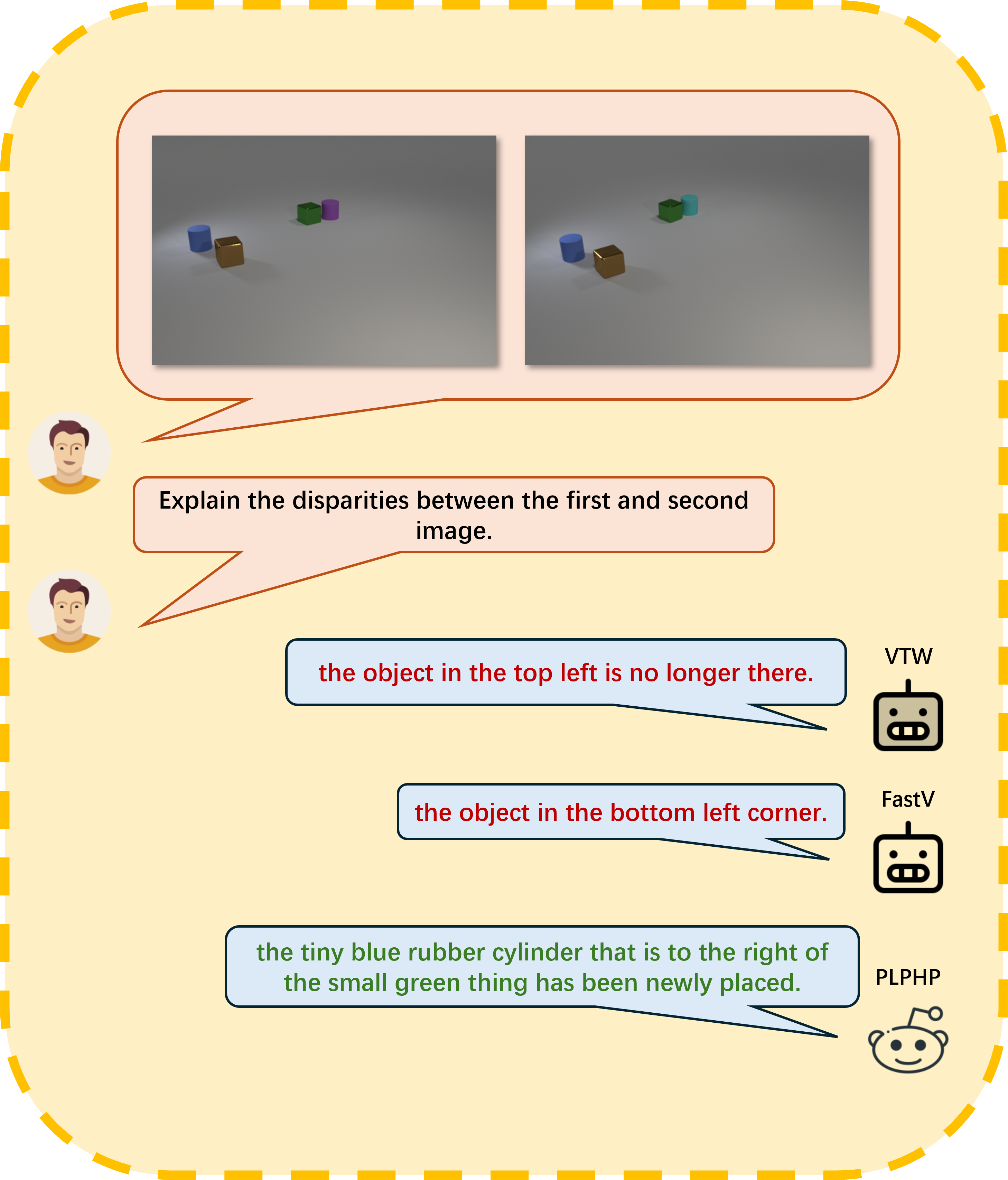}}
        \subfloat[]{
		\includegraphics[width=0.48\textwidth]{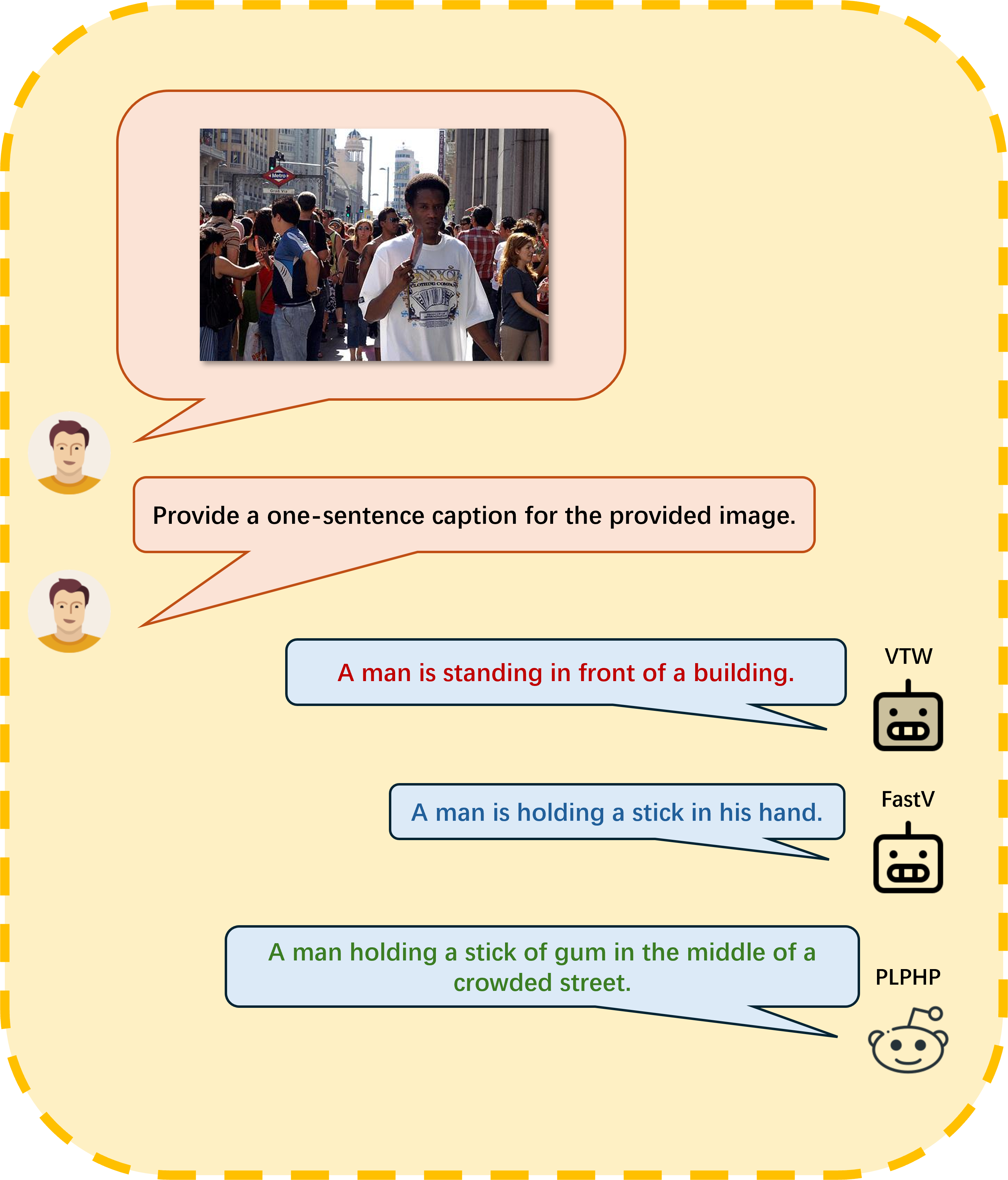}}
        \\ \quad \\ \quad \\
	\subfloat[]{
		\includegraphics[width=0.48\textwidth]{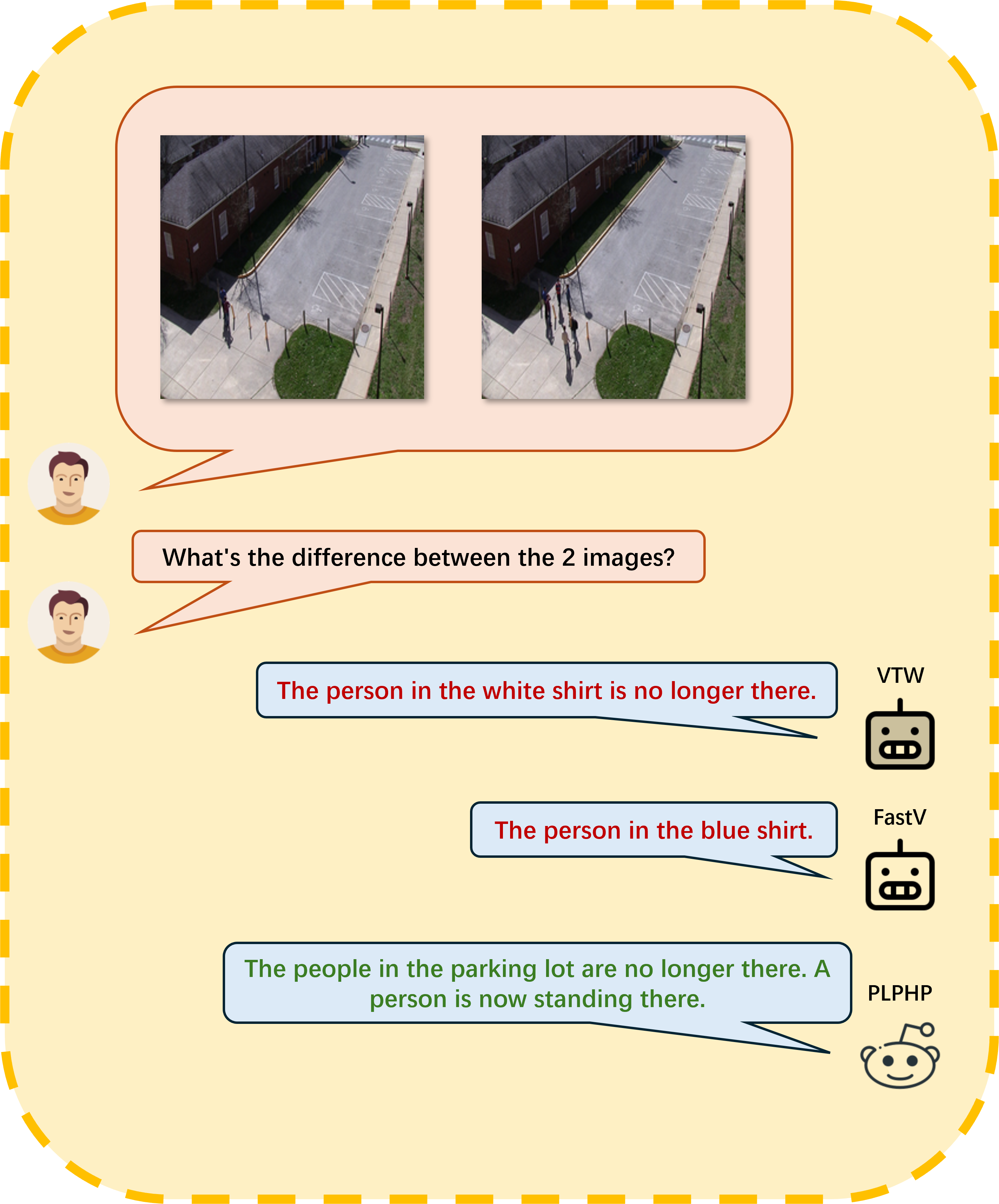}}
        \subfloat[]{
		\includegraphics[width=0.48\textwidth]{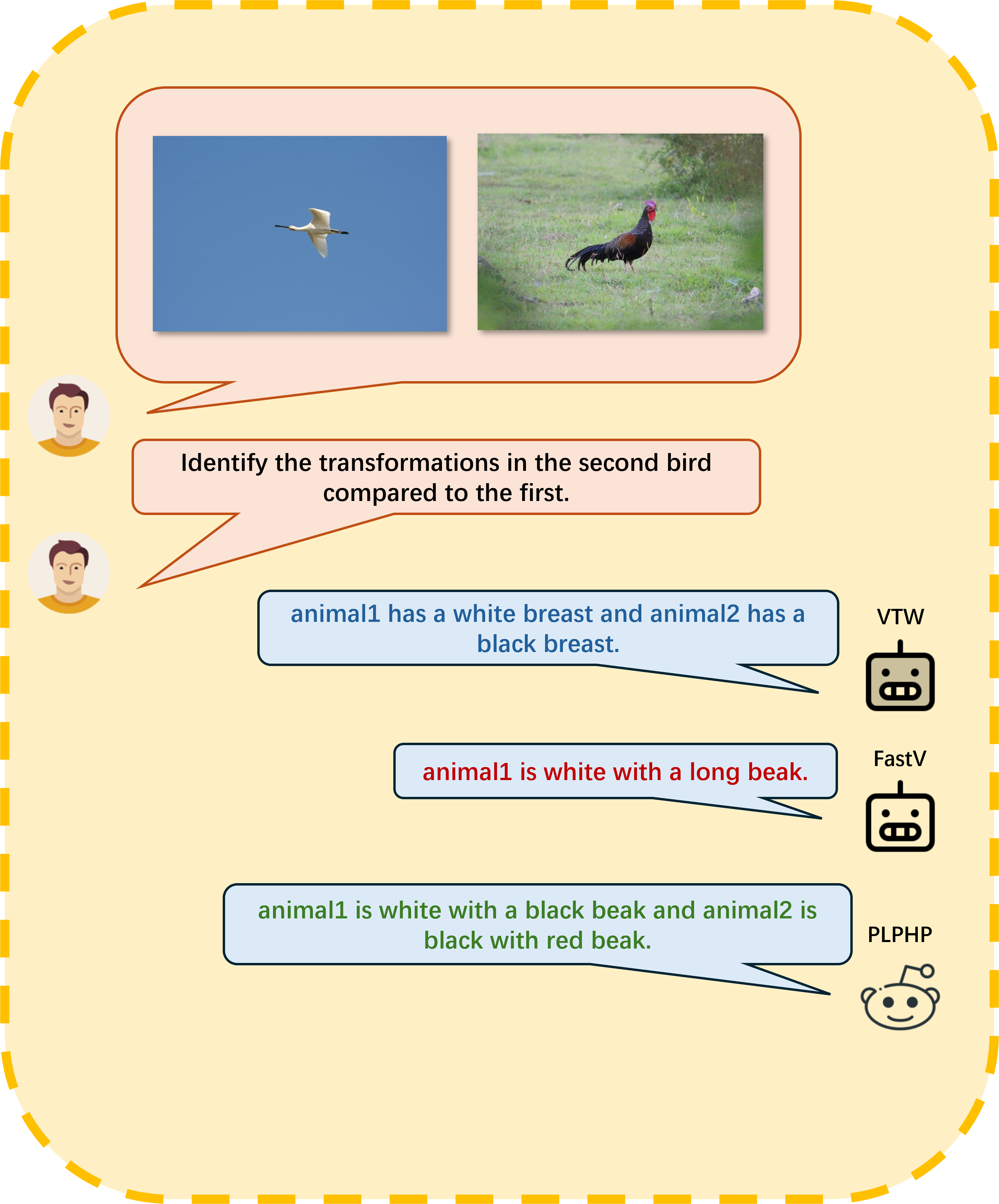}}
 %        \\
	% \subfloat[]{
	% 	\includegraphics[width=0.9\textwidth]{figs/appendix-case5.png}}
	\caption{\textbf{Multimodal Chatbots with different pruning methods.}}
		\label{fig:case-studies}
\end{figure*}

To showcase the effectiveness of our proposed method, we present a series of case studies in the form of multimodal chatbots, as shown in Figure \ref{fig:case-studies}.

\end{document}